\documentclass[sn-mathphys,Numbered]{sn-jnl}

\usepackage{enumitem}

\usepackage[utf8]{inputenc}
\usepackage[english]{babel}
\usepackage{amsmath}
\usepackage{graphicx}
\usepackage{amssymb}
\usepackage[table]{xcolor}

\usepackage{caption}
\usepackage{subcaption}
\usepackage{textcomp}
\usepackage{algorithm}
\usepackage{algpseudocode}
\usepackage{graphicx}%
\usepackage{multirow}%
\usepackage{amsmath,amssymb,amsfonts}%
\usepackage{amsthm}%
\usepackage{mathrsfs}%
\usepackage[title]{appendix}%
\usepackage{xcolor}%
\usepackage{textcomp}%
\usepackage{manyfoot}%
\usepackage{booktabs}%
\usepackage{algorithm}%
\usepackage{algorithmicx}%
\usepackage{algpseudocode}%
\usepackage{listings}%
\usepackage{geometry}
\theoremstyle{thmstyleone}%
\theoremstyle{thmstyletwo}%
\newtheorem{remark}{Remark}%

\theoremstyle{thmstylethree}%
\begin{document}

\title[Title]{Approximating $G(t)/GI/1$ queues with deep learning}


\author*[2,1]{\fnm{Eliran} \sur{Sherzer}}\email{eliransh@ariel.ac.il}

\author[1]{\fnm{Opher} \sur{Baron}}\email{opher.baron@rotman.utoronto.ca}

\author[1]{\fnm{Dmitry} \sur{Krass}}\email{dmitry.krass@rotman.utoronto.ca}

\author[3,1]{\fnm{Yehezkel} \sur{Resheff}}\email{Hezi.Resheff@gmail.com}

\affil*[1]{\orgdiv{Rotman school of management}, \orgname{University of Toronto}, \orgaddress{\street{105 St George St}, \city{Toronto}, \postcode{,  ON M5S 3E6, Canada}, \state{Ontario}, \country{Canada}}}

\affil[2]{\orgdiv{Industrial Engineering}, \orgname{Ariel University}, \orgaddress{\street{Ramat HaGolan St 65}, \city{Ariel}, \postcode{4070000}, \country{Israel}}}

\affil[3]{\orgdiv{The Hebrew University Business School}, \orgname{Hebrew University}, \orgaddress{\street{Shomrei Ha'har St}, \city{Jerusalem}, \postcode{9190501} \country{Israel}}}


\abstract{Many real-world queueing systems exhibit a time-dependent arrival process and can be modeled as a $G(t)/GI/1$ queue. Despite its wide applicability, little can be derived analytically about this system, particularly its transient behavior.  Yet, many services operate on a schedule where the system is empty at the beginning and end of each day; thus, such systems are unlikely to enter a steady state.  In this paper, we apply a supervised machine learning approach to solve a fundamental problem in queueing theory: estimating the transient distribution of the number in the system for a $G(t)/GI/1$. 

We develop a neural network mechanism that provides a fast and accurate predictor of these distributions for moderate horizon lengths and practical settings. It is based on using a Recurrent Neural Network (RNN) architecture based on the first several moments of the time-dependant inter-arrival and the stationary service time distributions; we call it the Moment-Based Recurrent Neural Network (RNN) method (\textit{MBRNN}).  Our empirical study suggests MBRNN requires only the first four inter-arrival and service time moments. 


We use simulation to generate a substantial training dataset and present a thorough performance evaluation to examine the accuracy of our method using two different test sets. We perform sensitivity analysis over different ranges of Squared Coefficient of Variation (SCV) of the inter-arrival and service time distribution and average utilization level. We show that even under the configuration with the worst performance errors, the mean number of customers over the entire timeline has an error of less than 3\%. We further show that our method outperforms fluid and diffusion approximations. 

While simulation modeling can achieve high accuracy (in fact, we use it as the ground truth), the advantage of the \textit{MBRNN} over simulation is runtime. While the runtime of an accurate simulation of a $G(t)/GI/1$ queue can be measured in hours, the \textit{MBRNN} analyzes hundreds of systems within a fraction of a second. We demonstrate the benefit of this runtime speed when our model is used as a building block in optimizing the service capacity for a given time-dependent arrival process. 

This paper focuses on a $G(t)/GI/1$,  however the \textit{MBRNN} approach demonstrated here can be extended to other queueing systems, as the training data labeling is based on simulations (which can be applied to more complex systems) and the training is based on deep learning, which can capture very complex time sequence tasks. In summary, the \textit{MBRNN} has the potential to revolutionize our ability for transient analysis of queueing systems.
 }




\maketitle

\section{Introduction} \label{sec:intro}

Time dependency is ubiquitous in practical queueing systems due to seasonality and weekly, daily, or hourly fluctuations in customer arrival patterns. Examples include time-dependent arrivals of calls to an inbound call center, patients to a hospital emergency department, planes to air traffic control, and trucks to container terminals. Other systems have cyclic or periodic arrivals, such as message volumes in IT systems or coffee-shop customers. In the language of queueing theory, such queues have time-dependent arrival parameters. In fact, it is harder to list applications with a fixed arrival rate over a sufficiently long horizon.

In many applications, service systems are empty at the beginning and end of each work day, and face time-varying arrival rates during operating hours; thus, the steady-state behavior, which is the focus of most analytical queuing methods, may never set in.  It is known that arrival dynamics can substantially impact the queueing system's performance~\cite{HWANG2004129}, and thus, must be considered in the design and control of such systems.

Despite their importance, accurate and fast solutions for modeling time-dependent behavior of queueing systems are sparse. Numerical or analytical methods are only available under restrictive assumptions. For example, the $M(t)/M(t)/1$ queue can be solved numerically via the Chapman–Kolmogorov Equations (CKEs)  using the Euler method or a Runge Kutta scheme~\cite{doi:10.1287/opre.23.6.1045}. 
However, even in this relatively simple Markovian system, the solution requires long computation times~\cite{doi:10.1287/ijoc.1050.0157}. While several approximation methods for transient systems have been proposed, for example, \cite{newell_1968_1,newell_1968_2, newell_1968_3, 1146391}, these methods suffer from limited accuracy (see Sections~\ref{sec:lit_rev} and ~\ref{sec:diff_comper} for further discussion and results).  

The most practical method for analyzing time-dependent transient queues is system simulation. While, given enough time, simulation can result in very accurate solutions (in fact, we use it to build our training library), it can be extremely time-consuming, in some cases prohibitively so. As discussed in Section~\ref{sec:simulation}, for each training instance, it took us nearly $160$ mins to generate the distributions of the number in the system in each time period. This makes simulation impractical for settings where near-real-time response is required and/or for an iterative decision process, possibly with a human in the loop. 
For examples, management may want to analyse system sensitivity to  various demand scenarios, requiring answers to multiple "what if" questions. Another example involves optimization of system capacity (processing rate) under time-dependent arrivals pattern and/or time-dependent service-level cosntraints, where we wish to ensure sufficiently low waiting time in each time period (see our example of optimization application in Section~\ref{sec:numeric}). 

To help overcome these limitations, we propose Moment-Based Recurrent Neural Network (MBRNN) - a novel machine learning-based method for approximating the number of customers in the system as a function of time. 

Our methodology is based on Recurrent Neural Networks (RNNs)~\cite{6789445}. RNNs are a class of artificial neural networks characterized by connections between nodes in the computation graph that form a cycle, allowing outputs from some nodes to affect the computations based on subsequent inputs to the same nodes. These connections allow RNNs to capture and exhibit temporal dynamic behavior. This feature creates a natural environment to analyze the queue state over time in the transient context. 

To take advantage of the attractive features of RNNs in our context we must represent a time-varying general arrival process as a discrete series of inputs (since RNNs represent time as a series of discrete intervals). To this end, we leverage insights obtained in applying Neural Networks to obtain the stationary occupancy distribution in a $GI/GI/1$ queue in our previous work~\cite{Sherzer23}, where we demonstrate that representing general arrival and service processes with a finite number of moments leads to extremely accurate estimates of steady-state queuing dynamics.  We thus use a finite-moment representation of the arrival process in each of the discrete-time periods, as well as a similar representation of the service process, resulting in the MBRNN algorithm.

We seek to infer the probability distribution $P(t)$ of the number of customers in the system in each period $t\in \{1, 2,...,T \}$. Our inputs are a very general time-dependent arrival process, a stationary service process, and the initial distribution of the number of customers in the system $P(0)$. The arrival process is assumed to be stationary within each time period but can change between periods, while the service process is assumed to be stationary over all periods (this assumption can be easily relaxed with only a minor change in the methodology but would require a larger training set). The arrival process in each time period is represented by the first four moments of the inter-arrival distribution (our analysis in Section \ref{sec:mom_analysis} shows that a higher number of moments does not lead to higher accuracy); the service process has a similar representation. The length of each period is scaled to equal the average service duration, allowing for a very granular representation of the transient behavior of the system. The main advantages of our methodology are prediction accuracy and speed. As demonstrated in \ref{sec:numeric}, our trained model can obtain predictions that are an order of magnitude more accurate than the alternative methodology (namely, the fluid approximation), with computation times of only fractions of a second versus over an hour required by simulation methods. To the best of our knowledge, this makes our approach the only practical technique for estimating the transient behavior of a general $G(t)/GI/1$ queue in ``real-time''. 

Our model is trained with $T=60$. However, as shown in Section \ref{sec:numeric}, the trained model can be applied to obtain predictions over much longer time periods (we demonstrate that the accuracy remains reasonably high for predictions over up to $180$ periods).  Moreover, the value of $T$ can be increased at the cost of generating a new set of training instances.

Our approach belongs to the general class of supervised machine learning models, meaning that the model is trained on a ``labeled'' set of instances, i.e., a set of $G(t)/G/1$ queues with known values of $P(t)$ for every $t=1, \ldots, T$. Since a trained model should be able to predict $P(t)$ for general inputs (not limited to the training set), the generation of training data is a major challenge in this work.  Specifically, we need to generate a large set of training instances that are sufficiently diverse with respect to inputs (i.e., time-varying arrival process and the service process) to enable the model to make accurate inferences for any provided input.   

To tackle the challenge of producing a collection of training instances that represent a wide variety of possible inputs, we
extend the approach of~\cite{Sherzer23} who used the  Phase-Type (PH) family of distributions, which is known to be dense with respect to all non-negative distributions (i.e., any given non-negative distribution can be approximated by a PH distribution to a pre-specified level of accuracy, see~\citep[Theorem 4.2, Chapter III] {Asmussen2003}). \cite{Sherzer23} developed an algorithm to 
generate PH instances that are very diverse and enabled their $GI/GI/1$ model to very accurately predict steady-state occupancy distributions for a wide variety of data sets.  We employ the same algorithm to generate different inter-arrival processes that are ``stitched together'' as described in Section \ref{sec:Training_Data} to generate time-varying arrival processes exhibiting a wide range of behaviors.  We also use the sample algorithm to generate the service process.  We apply simulation modeling to the resulting instances to compute $P(t), t=1, \ldots, 60$ distributions, thus creating ``labels'' for our training data.  

The trained model was applied to a variety of data sets, and its accuracy was compared to fluid approximation, as well as the diffusion approximation (the latter comparison is limited to the transient $GI/GI/1$ case since the diffusion approximation is not available for non-stationary arrival process suggested in this study). We note that the comparison was made over the mean number of customers and not the full distribution since only our method can make such approximations in this current setting.   MBRNN is shown to be at least 10x more accurate than the fluid approximation, generally more precise than the diffusion approximation, and much faster than any other method.  In particular, it is 1000x faster than simulation modeling - the only other methodology that can deliver a similar level of accuracy. 

While our development makes some seemingly restrictive assumptions (stationary service process, $T=60$ time periods), these are easy to relax with minimal changes to our methodology.  Creating sufficiently large training sets under less restrictive assumptions is mostly a matter of computational time.

To summarize, our main contributions are:
\begin{enumerate}[label=\roman*., itemsep=0pt, topsep=0pt]
    \item We show that an ML-based technique can accurately and nearly instantaneously approximate a fundamental queueing transient system, namely the $G(t)/GI/1$ queue while handling inter-arrival and service time distributions of great complexity. This technique is not limited to the $G(t)/GI/1$ but can also be extended to other systems, most directly to $G(t)/G(t)/1$. Furthermore, inference can be done instantly. 
    \item We develop an open-source package that infers the transient distribution of the number of customers in the system under $G(t)/GI/1$ (found in https://github.com/eliransher/MBRNN.git).
    \item We demonstrate how our method can be applied in fully data-driven settings (where moments of inter-arrival and service distributions are not known and must be estimated from the data), as well as in optimization or sensitivity analysis settings, where many instances need to be solved to enable the analysis. 
    \item We conduct an empirical moment analysis, examining the effect of the $i^{th}$ moments of the inter-arrival and service time distribution on the number of customers distribution. Our results suggest that the first four moments almost fully determined the queue length's stochastic properties. 
\end{enumerate}  

Before leaving the current section, we note that while our methodology is based on RNNs, an alternative deep learning architecture is the Transformer model class introduced in ~\cite{NIPS2017_3f5ee243} and later adapted for time-series data ~\cite{zhou2021informer}. These models are widely used in various domains due to their effectiveness in handling sequential data. However, empirically, we did not obtain better results from the Transformer architecture than the RNNs, and we thus prefer the RNNs that have a simpler architecture. 

The rest of the paper is organized as follows. In Section~\ref{sec:lit_rev}, we present related work, reviewing both transient analysis models and ML queueing studies. In Section~\ref{sec:problem_formulation}, we clearly define the problem we solve via the \textit{\textit{MBRNN}}. In Section~\ref{sec:data}, we explain how we generate data to train our model. In Section~\ref{sec:deep}, we provide details on the \textit{\textit{MBRNN}} framework; we explain how the data is processed, the RNN architecture, and the loss function used in training. In Section~\ref{sec:exp_setting}, we present our experimental settings, followed by the results in Section~\ref{sec:result}. In Section~\ref{sec:numeric}, we apply our methods in two applications: an optimization problem and data-driven inference. 
Section~\ref{sec:Discussion} discusses some intuition behind the \textit{MBRNN}, presents possible extensions, and concludes the paper.

\section{Related Work}\label{sec:lit_rev}

Analyzing transient queueing models is generally challenging. Only very particular systems have known numerical and analytical solutions.


One such system is the $M(t)/M/c$, for which the Chapman-Kolmogorov Equations (CKEs) were introduced by Kolmogorov~\cite{Kolmogorov31}. Analytical solutions for these differential equations are available only in specific scenarios, such as when the number of servers $c=\infty$. Nevertheless, it's possible to derive numerical solutions through techniques like the Euler method or a Runge-Kutta scheme. When the system has an infinite waiting room, it leads to an infinite series of differential equations. \cite{doi:10.1287/opre.23.6.1045} proposed an approximation method for such systems by substituting the infinite waiting room with one that is adequately large yet finite. Numerical solutions of these equations have been employed to evaluate the performance of various systems: Koopman~\cite{doi:10.1287/opre.20.6.1089} for an $M(t)/M/1/K$ system, Bookbinder~\cite{doi:10.1080/03155986.1986.11732012} in the analysis of an $M(t)/M(t)/1/K$ system with two independent queues and a shared server, and Van As~\cite{1146409} for a multi-class $M(t)/M(t)/1/K/NPPrio$ system. Additionally, this numerical approach is utilized in performance assessments within optimization algorithms, as demonstrated by Parlar~\cite{doi:10.1080/00207728408926548} and Nozari~\cite{https://doi.org/10.1002/nav.3800320208}. A key advantage of using the numerical solution for CKEs is its ability to capture the complete time-dependent distribution of state probabilities, which facilitates the computation of relevant quantiles~\cite{INGOLFSSON2002585}. Nonetheless, the method is limited to Markovian systems and is often hindered by lengthy computation times~\cite{doi:10.1287/ijoc.1050.0157}.

Newell suggests the diffusion approximation for $G(t)/GI/1$ systems in his trio of works~\cite{newell_1968_1,newell_1968_2, newell_1968_3}, specifically addressing the scenario of a rush hour when the arrival rate temporarily surpasses the service rate before subsiding. Duda~\cite{1146391} extends the use of diffusion approximation to evaluate the transient performance of a $G(t)/G/1$ system, which is not confined to rush hour scenarios characterized by fluctuating arrival rates. This broader application makes Duda's results relevant to our model, prompting a comparative analysis of performance metrics between his model and ours, as detailed in Section~\ref{sec:diff_comper}. For additional insights into transient queueing analysis, readers are directed to the comprehensive survey in~\cite{SCHWARZ2016170}.

Next, we turn to machine learning methods to analyze queuing models. Several works have utilized neural networks in time-dependent queueing analysis. The first to do this were Du, Dai, and Trivedi~\cite{du2016recurrent}. They trained a model to generate the precise time interval or the exact distance between two events. This carries a great deal of information about the dynamics of the underlying systems. This information is used in cases with large volumes of event data, such as healthcare analytics, smart cities, and social network analysis. Their objective is to learn the arrival pattern to the queue.  

In~\cite{garbi2020learning}, Garbi, Incerto, and Tribastone propose a novel methodology based on RNNs where analytical performance models are automatically learned from a running system using execution traces. Their goal in this work was to discover the topology of a queueing network and the service demands. Finally, ~\cite{ojeda2021learning} provides a scalable algorithm for service time inference that scales to the large datasets of modern systems. They provide a first bridge between queuing system modeling and Generative Adversarial Networks (GANs). 

While all these studies utilize the ability of a neural network to capture time-dependent dynamics very successfully, they approximate the queueing model inputs. Still, they do not analyze the queueing dynamics (e.g., queue length and waiting times). 

Kudou and  Okuda~\cite{10226861} also used an RNN architecture to predict transient queueing measures. Unlike our model, they predict the next customer waiting time given the last inter-arrival length, the number of customers in the system, and the previous waiting time. The gap between the need for solid prediction in transient queueing and the current literature is summarized well in Chocron et al., \cite{9976883},  as they compared ML performance against analytic methods, they suggest that for complex transient systems ''\textit{more adaptive ML algorithms, such as recurrent neural network (RNN) and specifically long short-term memory (LSTM), could provide better predictions.}"

Neural networks have been used for various other queue analysis tasks. In~\cite{Nii20} the stationary $GI/GI/c$ queue is analyzed. \cite{Kyritsis19} utilized a NN to predict waiting times. However, unlike our model or in~\cite{Nii20}, they focus on an online prediction setting conditioned on the current state of a network rather than on the stationary analysis of the system.  In a similar work~\cite{Hijry_2020}, evaluate the expected waiting times in an emergency room using deep learning. Finally,~\cite{wei2023sample} proposes an efficient Reinforcement Learning method that accelerates learning by generating
augmented data samples. The proposed algorithm is data-driven and learns the
policy from data samples from both real and augmented samples. This method
significantly improves learning by reducing the sample complexity so that the dataset only needs sufficient coverage of the stochastic states. The application of their work is finding optimal or near-optimal control/scheduling policy.

\section{The Inference Problem, Input, and Solution Approach}\label{sec:problem_formulation}
We define the inference problem in $G(t)/GI/1$ queues, discuss the input data required to model the problem, and provide a roadmap of our machine learning solution approach that is further described in Section \ref{sec:deep}.

\subsection{The Inference Problem}
Consider a $G(t)/GI/1$ system. We restrict our discussion to focus on finding the probabilities of having $0 \leq i\leq l$  customers in the system at each $(t=1,...,T)$. We note, however, that the general method developed in this paper is suitable in principle for inference of any performance measure of interest. The cutoff value $l$ choice reflects the maximum number of customers in the system at any given time and is discussed further in Section~\ref{sec:simulation}.  

For simplicity, we choose a unit of time such that the average service time equals a one-time unit. This choice determines how frequently the process is monitored. Most practical single-server systems do not have large changes in their dynamics faster than during a service time. Thus, this choice seemed sufficient. Let  $T$ represent the total horizon length we wish to infer. Due to the discrete nature of RNNs, we infer the distributions at the end of each of the $T$ periods. Let $\textbf{P}(t)$ be an $l+1$ size vector, where the $i^{th}$ element $P_i(t)$ denotes the probability of having $i$ customers in the system at time $t$. Thus, we are interested in infereing the sequence $\{\textbf{P}(1), \textbf{P}(2),...,\textbf{P}(T) \}$.  

\subsection{The Input}
The input to the model is sequential. At each time point, this input 
consists of time-independent and time-dependent data. The discussion of the time-dependent data is a reflection of the $G(t)$ processes we consider in our inference problem.

\subsubsection{Time-independent Data }
The constant data input includes: (i) $\textbf{S}$ is a  $n_{service}$ size vector, which represents the first $n_{service}$  moments of the service time distribution.  (ii)  the system's initial state, i.e., the distribution of the number of customers in the system at time $t_0$ (for $0 \leq i\leq l$). We denote the initial state by $\textbf{P}(0)$. 

\subsubsection{Time-dependent Data}\label{sec:arrival_proc}
The $G(t)$ arrival process is time-dependent, and so is the input data to characterize it. We restrict our attention to arrival processes defined as follows. During each period, $j\in \{1, 2,...,T \}$, i.e., during $t \in (j-1,j]$ a renewal arrival process is governed by a continuous non-negative inter-arrival time distribution, with finite first $n_{arrival}$ moments. Let $\textbf{A}(j)$ be an $n_{arrival}$ size vector for the first $n_{arrival}$ inter-arrival moments at the $j^{th}$ period; thus  $\textbf{A}(j)$ is related to $t \in (j-1,j]$.

\subsubsection{Training Data}\label{sec:Training_Data}
\begin{figure}
\centering
\includegraphics[scale=0.55]{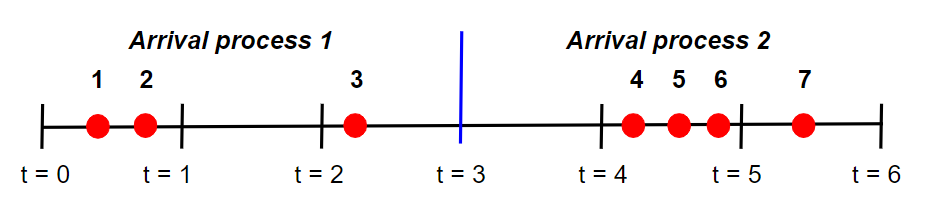}
\caption{An arrival process example.}
\label{fig:arrival_proc}
\end{figure}

The training data for the model was generated as follows. For each system we divide the fixed horizon $T$ into $m \in [1,...T]$, not necessarily identical, intervals of an integer number of periods. The length of each of the $m$ arrival processes is such that their total length is $T$. During each such interval, the arrival process is stationary. An example is shown in  Figure~\ref{fig:arrival_proc} where the first interval (namely \textit{Arrival process 1}) is of length 3 then $\textbf{A}(1)=\textbf{A}(2)=\textbf{A}(3)$, i.e., the $n_{arrival}$ inter-arrival moments at the $j=1,2,3$ periods
are identical. The renewal arrival process changes when we switch from one interval to another. This change has two implications. First, the $\textbf{A}(j)$ changes, and second, the new arrival process starts afresh. That is, the residual of the previous arrival process is no longer relevant. In Figure~\ref{fig:arrival_proc}, the second interval (namely, \textit{Arrival process 2}) starts at period 4 and is also of length 3. Then $\textbf{A}(3) \neq \textbf{A}(4)=\textbf{A}(5)=\textbf{A}(6)$. Moreover, the renewal arrival process of this second interval starts afresh at $t=3$.


Figure~\ref{fig:arrival_proc} shows an example arrival process generated in this way. The red dots over the timeline are the arrivals numbered from 1 to 7, and the blue line, at $t=3$, partitions the timeline of length $T$ into the two arrival processes. \textit{Arrival process 1} started at $t=0$ and ended at $t=3$. At $t=3$, \textit{Arrival process 2}  starts afresh, and ends at $t = 6$ (which is immaterial, as $T=6$). The age of the inter-arrival time at the beginning of each arrival process (i.e., at $t=0$ and $t=3$) is 0. In the example above, once the $3^{rd}$ customer arrives, a new inter-arrival is generated from \textit{Arrival process 1}. Still, this arrival never occurred since we reached $t=3$ before its arrival time. The new \textit{Arrival process 2} starts afresh at $t=3$, and the arrival of the $4^{th}$ customer follows \textit{Arrival process 2}. As we can see, when switching from one time period to another, it can either mean we remain in the same arrival process (e.g., from the second to the third) or initiate a new arrival process (e.g., from the third to the fourth).

As we set $T\in \mathbb{N}$ and $n=T$ in our data experiments, i.e., there are $n$ time intervals of size 1; we wish to predict the sequence $\{\textbf{P}(1), \textbf{P}(2),...,\textbf{P}(T) \}\in [0,1]^{l\times T}$, via the input $\{\textbf{A}(1), \textbf{A}(2),...,\textbf{A}(T) \}$, $\textbf{S}$ and $\textbf{P}(0)$.

\subsection{\textit{MBRNN}: Moment Based RNN}
\label{sec:overview}

\begin{figure}
\centering
\includegraphics[scale=0.4]{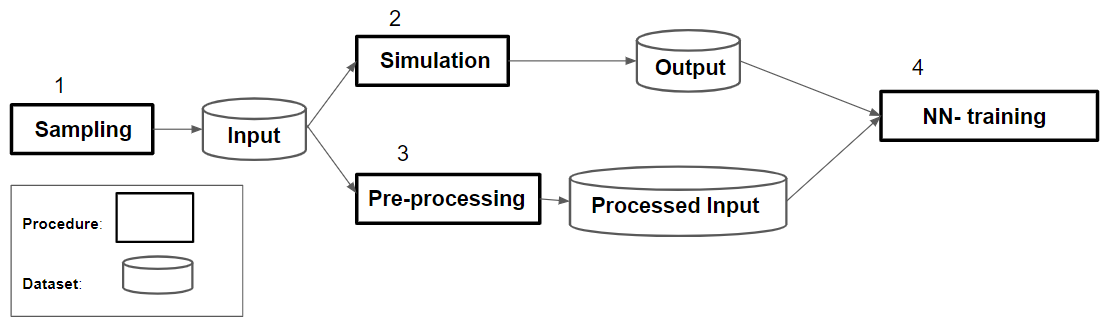}
\caption{Work-flow diagram of our learning procedure. }
\label{fig:diagram_overview}
\end{figure}

\begin{figure}
\centering
\includegraphics[scale=0.55]{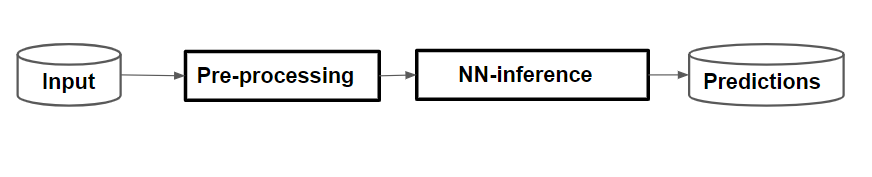}
\caption{Work-flow diagram for inference. }
\label{fig:diagram_inf}
\end{figure}

The main phases of our solution are presented in~Figure~\ref{fig:diagram_overview}, which outlines the learning procedure, and Figure~\ref{fig:diagram_inf}, which presents the inference process. 

Steps 1-3 in~Figure~\ref{fig:diagram_overview} represent the training input and output data generation, as detailed in Section \ref{sec:data}. We start by \textit{generating input samples}, step 1 (described in Sections \ref{sec:dists_sampling} and \ref{sec:arrival_pattern_sampling}).  This step results in the \textit{Input}. In step 2, \textit{Simulation}, we compute the distribution of the number of customers in the system using a discrete event time simulation, described in Section~\ref{sec:simulation}. 
The \textit{Output} from the simulation, i.e., the probabilities of having $0,...,l$ customers in the system for every $t=1,...,T$, is later used as the target when learning from the training data.  

The inter-arrival and service time distributions are \textit{Pre-processed} in step 3, where we compute the first $n_{arrival}$ and $n_{service}$ moments, respectively. Note that the hyper-parameters\footnote{Hyper-parameters are different factors of the models which are not the RNN weights. Along with $n_{arrival}$ and $n_{service}$ moments, it can be the number of hidden layers or nodes within a hidden layer. } $n_{arrival}$ and $n_{service}$ require tuning. We then standardize the computed moments, as detailed in Section \ref{sec:intput preprocessing}.
The processed input and training outputs are fed into the NN for \textit{training} in step 4, (as described in Section \ref{sec:network}), and then training takes place.  Once the training procedure is done, we can make inferences according to the diagram in Figure~\ref{fig:diagram_inf}. Namely, we pre-process the input as $\{\textbf{A}(1), \textbf{A}(2),...,\textbf{A}(T) \}$, $\textbf{S}$ and $\textbf{P}(0)$ and then apply the NN for inference.

\section{Data Generation}\label{sec:data}

We generate training data in three phases. Phase I: The creation of a comprehensive inter-arrival and service times distribution library. At this phase, we wish to create a versatile distribution library that covers a wide range of distributions likely to include most real-world situations. Phase II: Generating the arrival rate pattern and the system's initial state. At this phase, we wish to cover a wide range of arrival patterns. Phase III: simulate these systems.

To present the data generation process, we define the following: for a given time period, let $\rho_i$ be the \textit{period utilization level} be the period arrival rate times the service mean value. We note that this measure is not equivalent to the asymptotic proportion in which the server is busy, as it also depends on the number of customers at the beginning of the period. However, it allows control of the overall stability of the queue. Since the mean service time is stationary, the arrival pattern determines the period utilization level. 

In phase II, we also sample an initial state for the system, which refers to the number of customers in the queue at $t=0$. Once phases I and II are done, we have full information about the different systems sampled, leading us to Phase III, where we simulate these systems and keep track of the distribution of the number of people in the system at each $t=1,...,T$.
We note that in our data generation process, we use $T=60$, equivalent to an hour of real-time when the average service time is one minute. Next, we describe the process in detail.


\subsection{Phase I: Sampling inter-arrival and service time distributions}\label{sec:dists_sampling}

We use the state-of-the-art sampling technique proposed in~\cite{Sherzer23}.  To sample distributions we utilize the Phase-Type (PH) family of distributions. As mentioned above, this family is known to be dense within the class of all non-negative distributions. While we sample from PH distributions of a constant degree, this property still allows us, in practice, to achieve a diverse training set. Indeed, in~\cite{Sherzer23} a wide distribution coverage is demonstrated and favorably compared with alternative sampling methods.

We sample distributions with a mean value of 1. For service time distributions, we assume, without loss of generality, that the mean service time is 1. Consequently, the arrival rate and period utilization levels coincide, simplifying our analysis. For the inter-arrival distribution, the sampled distributions from the library will be scaled according to the arrival rates sampled in Phase II to allow different arrival rates. The Squared Coefficient of Variation (SCV) of the sampled distribution used in this study ranges from 0.0025 to 106.46, demonstrating a versatile set of distributions. 
 
\begin{remark}
Recall from Section~\ref{sec:problem_formulation}, inference is made for each $t=1,...,T$. This means a single period length equals the mean service time. As such, the time granularity of our inference allows us to monitor the queue very closely and quickly pick up any changes.
\end{remark}

\subsection{Phase II: sampling the arrival rate pattern and initial state} \label{sec:arrival_pattern_sampling}
We sample $\{\textbf{A}(1), \textbf{A}(2),...,\textbf{A}(T) \}$, $\textbf{S}$  and $\textbf{P}(0)$. We commence with $\textbf{P}(0)$. We allow up to $k<l$ customers at $t=0$. We sample uniformly over the $k$ simplex. In our settings,  $k=30$.


$\textbf{S}$ and the set $\{\textbf{A}(1), \textbf{A}(2),...,\textbf{A}(T) \}$ are sampled via the library generated in Phase I. The sampling procedure for $\textbf{S}$ is straightforward as we need to sample only one distribution (since it does not vary with time), and the mean value remains 1; hence, no scaling is required. However, sampling the sequence $\{\textbf{A}(1), \textbf{A}(2),...,\textbf{A}(T) \}$ is more involved as we explain next. 

For this purpose, we next discuss several properties of the queue dynamics we wish to capture with respect to the arrival rate pattern.

\noindent (\textbf{i}) We wish to include a cyclic arrival pattern as is often the case in many service systems (e.g.,~\cite{doi:10.1287/msom.2013.0474}). For this purpose, we sample a \textit{CycleLength}, i.e., the number of periods within a cycle, and then concatenate all cycles until we reach $T$, while allowing the last cycle to be truncated accordingly so the total time will sum to $T$. 

\noindent  (\textbf{ii}) Let $\bar{\rho}$ be the cycle utilization levels, which is the average of the  $\rho_i$ values for $i \leq CycleLength$, for all over a single cycle as defined below in Equations~\eqref{eq:cycle_util} and~\eqref{eq:rate_equals_util}.  We wish to keep $\bar{\rho}$ in a single cycle within a reasonable range. Being more concrete, we focus on $0.5 \leq \bar{\rho} \leq 1$. We do not focus on low $\bar{\rho}$ values where the queue dynamics are less complex and interesting. Although $\bar{\rho}<1$, $\rho_i$ may exceed 1 for $i \geq 1$. By bounding the $\bar{\rho}$ at 1, we desire richer queue dynamics, where the system can seldom empty itself. Recall from above that $\rho_i$ is not it is not equivalent to the proportion of time in which the server is busy within a period. Consequently, $\bar{\rho}$ is not equivalent to the proportion of time in which the server is busy within a cycle. 

\noindent (\textbf{iii}) we wish to keep the same inter-arrival distribution for successive periods. Thus, we do not restrict our method to different inter-arrival distributions every period. Let $m$ denote the number of different inter-arrival distributions out of \textit{CycleLength} periods, where $m \leq \textit{CycleLength}$. 

We next describe the procedure of sampling $\{\textbf{A}(1), \textbf{A}(2),...,\textbf{A}(T) \}$. We first sample $\textit{Cycle Length} \sim U(T/3, 2T/3)$, hence enabling a cyclic arrival pattern\footnote{ If $\textit{Cycle Length} = T/3 $ there will be three full cycles, and if $\textit{Cycle Length} = 2T/3$ then we get one full cycle followed by a partial cycle. }. Then, we sample $m \sim (1, Cycle Length)$. Let $\lambda_i$ be the arrival rate (and hence $\rho_i$ since service mean value is 1) at the $i^{th}$ period.  For each $i=1,...,T$,  $\lambda_i \sim U(0.5,10)$, the arrival rate may differ from one another up to 20-fold. Let $\bar{\lambda}$ be the average arrival rate of a single cycle, where  

\begin{align}\label{eq:cycle_util}
    \bar{\lambda} = \sum_{i=1}^{CycleLength}\lambda_i/ Cycle Length
\end{align}

Since the service rate is constant at 1, our sampling procedure requires 

\begin{align}\label{eq:rate_equals_util}
\bar{\lambda}=\bar{\rho}  , \thinspace \lambda_i = \rho_i \thinspace \forall i \leq Cycle Length  
\end{align}

However, since   $\lambda_i$ for $ i \leq CycleLength$ are sampled independently from a uniform distribution, we must enforce this requirement. For this purpose, we apply the following normalization, $\lambda_i 	\leftarrow \bar{\rho}(\lambda_i/\bar{\lambda})$ for $i\leq CycleLength$.

\begin{figure}
\centering
\includegraphics[scale=0.55]{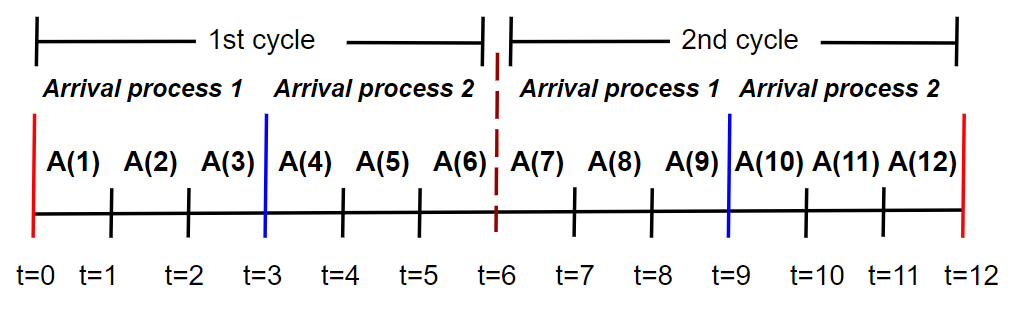}
\caption{Input of arrival process.}
\label{fig:arrival_proc_input}
\end{figure}

To conclude, for each system, we have a total of $m$ in different arrival processes. As such, we sample $m$ different inter-arrival distributions (according to Part A) and scale them according to the arrival rate sampled here.  To demonstrate, we follow the example in Figure~\ref{fig:arrival_proc_input}. In this example: $T=12$,  $CycleLength=6$. Thus, we have two cycles of 6 periods. In this example, each cycle consists of two arrival processes of 3 periods, hence $m=2$.  This means that $\textbf{A}(1)=\textbf{A}(2)=\textbf{A}(3)=\textbf{A}(7)=\textbf{A}(8)=\textbf{A}(9)$, and 
$\textbf{A}(4)=\textbf{A}(5)=\textbf{A}(6)=\textbf{A}(10)=\textbf{A}(11)=\textbf{A}(12)$.

\begin{figure}[]
\centering
   \begin{subfigure}[b]{0.9\textwidth}
   \includegraphics[width=1.\textwidth]{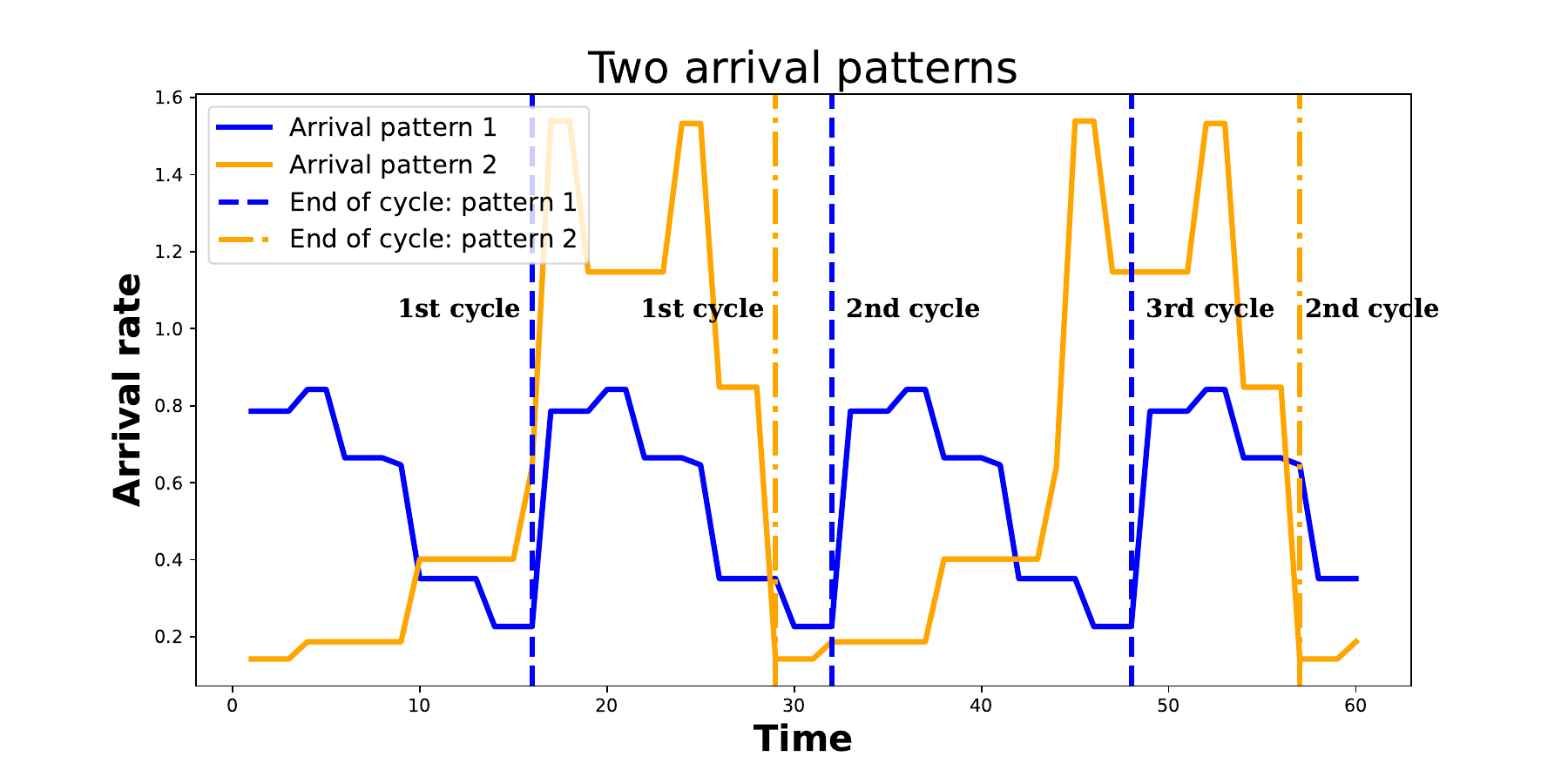}
   \caption{Two examples}
   \label{fig:arrival_pattern_2_examples} 
\end{subfigure}
\begin{subfigure}[b]{0.9\textwidth}
   \includegraphics[width=1.\textwidth]{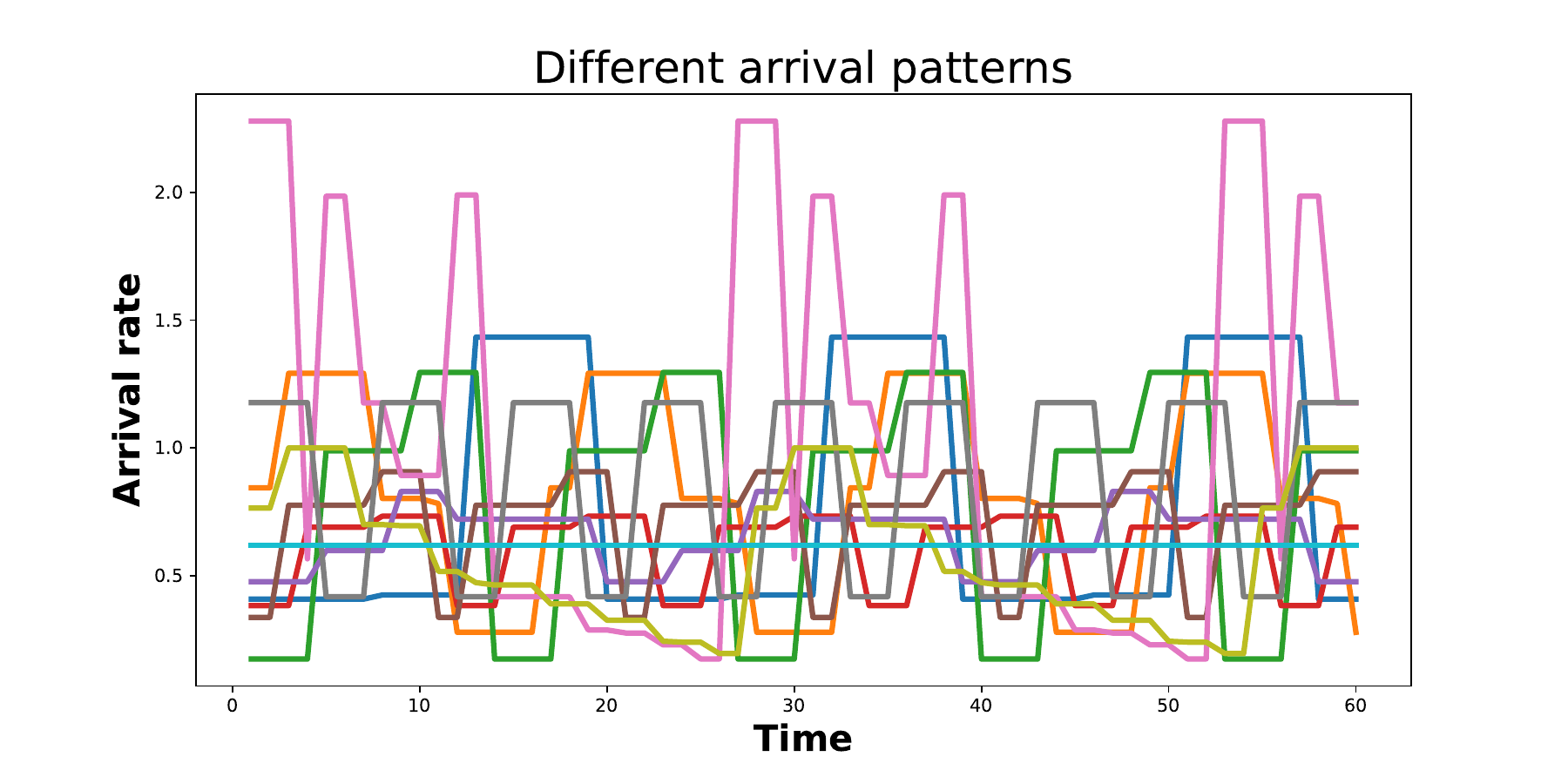}
   \caption{Ten examples}
   \label{fig:arrival_pattern_12_examples}
\end{subfigure}
\caption{Arrival pattern illustrations.}
\end{figure}

For illustration purposes, we present in Figures~\ref{fig:arrival_pattern_2_examples} and~\ref{fig:arrival_pattern_12_examples} arrival pattern examples. In Figure~\ref{fig:arrival_pattern_2_examples}, we have two arrival patterns that are quite different, with cycle lengths of 16 and 29. Arrival pattern 1 has an unimodal behavior and comes with a relatively smaller range of arrival rates. In contrast, arrival pattern 2 is more chaotic with a large range of rates (e.g., the largest rate is about 15-fold larger than the smallest). In~\ref{fig:arrival_pattern_12_examples}, we give ten examples to illustrate the wide range of possible ranges. We can have a constant rate for all time units but also fluctuate unstable rates.  

\subsection{Phase III: Simulation}\label{sec:simulation}

Labeling our training data is done via simulation. We first sample $\textbf{P}(0)$ for each run. Then, we generate arrivals and service times according to $\textbf{S}$ and the set $\{\textbf{A}(1), \textbf{A}(2),...,\textbf{A}(T) \}$. We keep track of the number of customers in the system at each period $t_i$ for $i=1,...,n$. For each setting of $\{\textbf{A}(1), \textbf{A}(2),...,\textbf{A}(T) \}$, $\textbf{S}$  and $\textbf{P}(0)$. We simulate $NumSim$ replication and extract the number in the system distribution. The value of $NumSim$ is discussed in Section~\ref{sec:runtimes}. 

To provide the computed distribution as input to the NN as a finite and fixed-dimensional vector, we truncate our computation at $l$ such that the probability of having more than $l$ customers is smaller than $\delta=10^{-10}$. Then, $l=50$ achieves the desired performance in all our generated samples. This corresponds to covering the probability of having between $0$ and $49$ customers in the system; the last value of the output vector is set to the probability of having $50$ or \textit{more} customers. 

The system is stable as long as the horizon $T=60$; we expect a total of less than $60$ customers. Intuitively, having $l=50$ or more customers in the system simultaneously at any given time would happen with an extremely low probability. This is the case in our simulations (even given that our initial state is distributed over $[0,30]$).

The final output is an $(T,l+1)=(60,51)$-dimensional matrix, where $T$ is the number of periods, and $l+1$ is the vector length of the number of customers in the system for a given sample and time period. 

\subsection{Training Set Runtimes and Accuracy}\label{sec:runtimes}

The value of $NumSim$ affects the trade-off between runtimes and the accuracy of the labels generated by the simulation.  The larger $NumSim$ is, the more accurate the labels, but the longer it takes to generate them. We run a total of 720k systems, each with NumSim=20000 replications, for $t \in [0,60]$, as our training set.
All of our numerical work, here and in Section 6,  were performed on one of the Beluga, Cedar, Graham, or Narval networks of Compute Canada.  The average run time for $20000$ replications is 159.37 minutes.

\begin{figure}
\centering
\includegraphics[scale = 0.6]{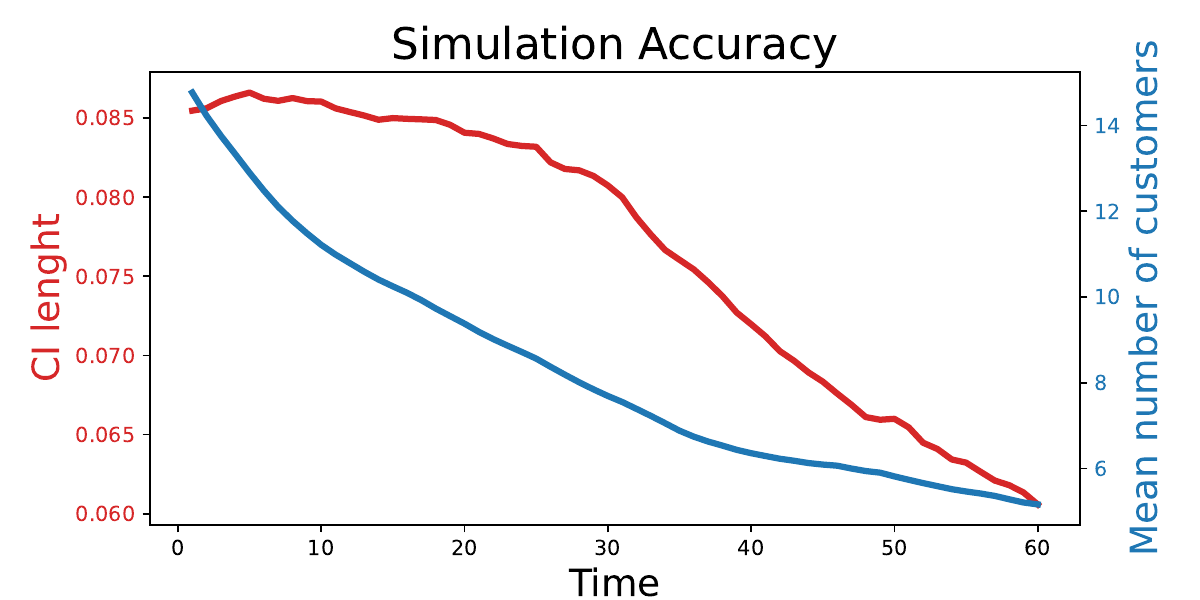}
\caption{Average length of confidence interval for the average number of customers in the system as a function of time. }
\label{fig:CI}
\end{figure}

We run the simulation for 300 $G(t)/GI/1$ to examine the labeling consistency. We simulate each queue ten times, each with $20000$ replications. We then measure the length of a 95\% confidence interval (CI) of the average number of customers in the system. We chose 20000 as the initial test showed that this led to a confidence interval lower than 0.1 for all time periods. Figure~\ref{fig:CI} represents the length of the CI (the red plot) as a function of time. As the figure demonstrates, the CI levels start high- at the beginning of the simulation and decrease with time. The high CI levels early in the simulation reflect the sizeable average number of customers in the system (as reflected in the blue plot).  As expected, the average initial number of customers is around 15. The negative drift depicted in the figure follows as even when this number is increasing for some queues, its average over 300 queues is decreasing as the overall utilization of each queue $\overline{\rho}<1$.

\section{\textit{MBRNN} model}\label{sec:deep}
This section provides details of our \textit{MBRNN} model, including input pre-processing, the network architecture, and the loss function. 

\subsection{Input pre-processing}\label{sec:intput preprocessing}
As mentioned above the input for a single $G(t)/GI/1$ prediction problem is defined by  $\{\textbf{A}(1), \textbf{A}(2),...,\textbf{A}(T) \}$, $\textbf{S}$  and $\textbf{P}(0)$. 

We take the following steps to process this input into sequential data in the format of matrices, as this is the input our NN requires. $\textbf{P}(0)$ is inserted into to the \textit{MBRNN} without pre-processing. As mentioned above, since  $\{\textbf{A}(1), \textbf{A}(2),...,\textbf{A}(T) \}$, and $\textbf{S}$, are distributions, we   represent them by the first $n_{arrival}$ and $n_{service}$ moments. Similarly to~\cite{Sherzer23}, we apply a log transformation for all inter-arrival and service time moments to reduce the range of possible values. This aligns with the common practice in NN models to ensure that input values fall within a similar range. Note that as we increase the order of moments, their values increase exponentially; thus, even using standardized moments does not ensure range similarity (and adds to the computational burden). In our experience and as in~\cite{Sherzer23}, the $log$ transform is simpler to apply and works better to reduce the range of values of all inter-arrival and service time moments.

The input for $t^{th}$ period, which refers to the time between $t-1$ to $t$, for $t=1,...,T$, consists of three parts. Inter-arrival log moments, service time log moments, and the initial state at $t=0$, i.e., $\textbf{P}(0)$. Thus, the input for any period, for $t=1,...,T$, is a vector of size  $n_{arrival} + n_{service}+ |\textbf{P}(0)|$. Since we consider $T$ periods, in total, the input for a single data point is a $T$ by $(n_{arrival} + n_{service}+|\textbf{P}(0)|)$ matrix.


\subsection{\textit{MBRNN}}\label{sec:network}
In a so-called \textit{vanilla} RNN, if the sequences are long, the gradients (which are used for learning the model weights with gradient descent) computed during training (with the backpropagation algorithm), are either vanishing (multiplication of many matrices with singular values less than 1 in absolute value) or exploding (singular values larger than 1 in absolute value), either way causing model training to fail.

To overcome this problem, we use the Long Short-Term Memory (LSTM) architecture as our RNN model. This modified RNN architecture tackles the problem of vanishing and exploding gradients and addresses the
problem of training over long sequences and retaining memory \cite{hochreiter1997long}. All RNNs have feedback loops in the recurrent layer. The feedback loop retains information in “memory” over time. However, training standard RNNs to solve problems that require learning long-term temporal dependencies can be challenging. The LSTM approach is especially suited for our problem as the queue status heavily depends on previous states, e.g., the initial state. However, the effect of previous states weakens with time, and this needs to be considered.  

\begin{figure}
\centering
\includegraphics[scale = 0.35]{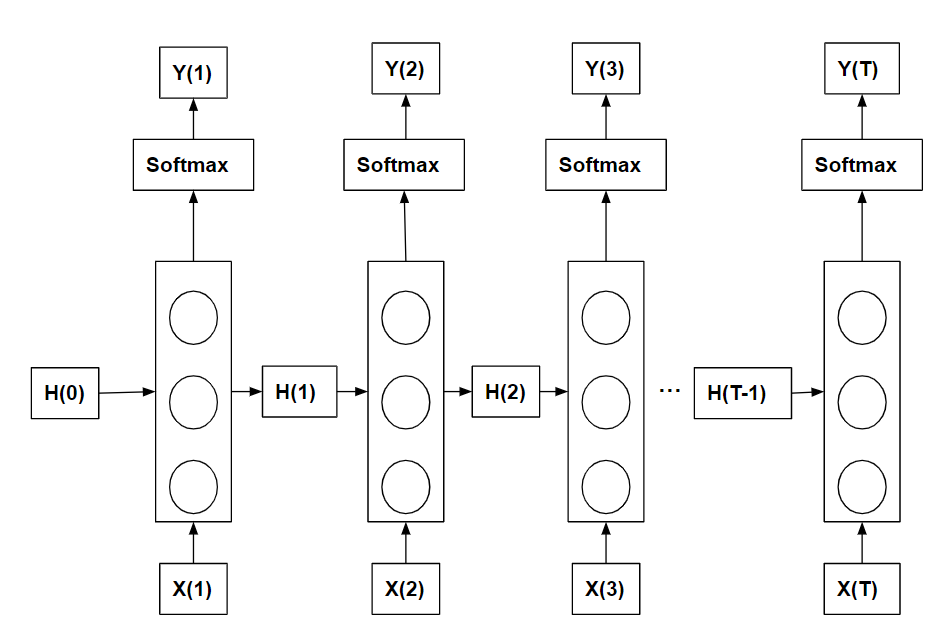}
\caption{Diagram of our RNN. $H(i)$ is the $i^{th}$ hidden input, $X(i)$ is the $i^{th}$ input and $Y(i)$ is the $i^{th}$ output. The rectangle with the three circles is the NN. }
\label{fig:NN_diagram}
\end{figure}

Figure~\ref{fig:NN_diagram} depicts the model architecture.  Since our output layer is a distribution, we applied the Softmax function\footnote{$SOFTMAX_i ([a_1,...,a_T]) = \frac{exp(a_i)}{\sum_{j=1}^n exp(a_j)}, i=1, \ldots, T$} as presented in Figure~\ref{fig:NN_diagram}, which guarantees that the output-layer sum up to 1. In the queueing context, for some time $t$, the NN must understand the current situation given the input $X(t)$, which consists of $A(t)$, $S$ and $P(0)$,  and the past information retained in $H(t-1)$, which stores the relevant history of the queue. While the RNN is trained over the first $T$ periods, there are no limits to the number of periods one can make inferences on. This is mainly due to the fact that all the time-step NNs in the architecture (the rectangles with the circles in Figure~\ref{fig:NN_diagram}) are identical. Essentially, each time step is computed by the same function of the input ($X$) and state ($H$). 


\subsection{Loss function} 
As is common in deep learning, we divide the training data into batches of size $B$ (and update the weights in the networks after each batch)~\cite{howard2020deep}. The hyper-parameter $B$ is selected during the finetuning process; see details  in~\ref{append:Fine-tuning}. As noted earlier, the output distributions for each time step of each training instance's system were truncated to size $l$. 
For a given batch, let $Y$ and $\hat{Y}$ represent a tensor of size $(B, T, l)$, of the true values of the transient distribution and the RNN predictions, respectively. This tensor is composed, therefore, of $B$ stacked matrices relating to each of the examples, each of size $T \times l$ that represent the distribution of customers in the system at each of the $T$ time-steps. 
Our loss function $Loss(\cdot)$ is given by

\begin{equation}  \label{eq:loss}
     Loss(Y,\hat{Y}) =   \frac{1}{B\cdot T}\sum_{i=1}^{B}\sum_{j=1}^{T}\sum_{k=0}^{l}|Y_{i,j,k}-\hat{Y}_{i,j,k}|+ 
     \frac{1}{B\cdot T}\sum_{i=1}^{B}\sum_{j=1}^{T}max_k(|Y_{i,j,k}-
     \hat{Y}_{i,j,k}|),
\end{equation}

\noindent where the values $Y_{i,j,k}$ and $\hat{Y}_{i,j,k}$ is the probability of having in  $k$ customers in the system for the  $i^{th}$ sample in the batch and the $j^{th}$ period for the ground truth and the predictions, respectively. As stated, the Hyper-parameter $B$ is tuned during training, where $l=50$ was set in our setting. 

The first term in our loss function Equation (\ref{eq:loss}) is the mean absolute distance between elements of the true and predicted distributions, while the second is the average maximum absolute error within each distribution. The rationale for the first term is that it offers an upper bound on the maximum difference between the two cumulative distribution functions (CDFs) for each sample. This concept is closely associated with the Wasserstein-1 distance, which is statistically justified when comparing distances between two distributions, see \cite{hallin2021}. The second term is inspired by common queueing applications, where it is generally preferable to make several small errors rather than a single large one. This is especially important because large errors typically occur in the tail, which is critical for maintaining service-level guarantees. Additionally, this term is closely related to the Kolmogorov-Smirnov two-sample test.

Note that both terms are equally weighted in our function. Although we could introduce an additional hyperparameter to adjust the relative importance of each term, the current form of the loss function seems sufficient, resulting in highly accurate predictions, as described in the following section.


\section{Experimental setting}\label{sec:exp_setting}
In this part, we describe the generated datasets used to train and validate our model, our training procedure, the metrics used to evaluate our model, and the tuning of hyperparameters for our final model.

\subsection{Datasets}
Below, we give details for our training and validation and test sets in Sections~\ref{sec:traninig_val_sets} and~\ref{sec:testsets}, respectively.

\subsubsection{Training and validation sets}\label{sec:traninig_val_sets}
To train the \textit{MBRNN} model, we require two different datasets: training, used to update the network parameters, and validation, used to tune the hyper-parameters, e.g., choose the hyper-parameter $B$~\cite{howard2020deep}, (see Section~\ref{sec:tunnig} for further details). We generated both using the procedures described in Section~\ref{sec:data} above. The training and validation sizes are 720,000 and 36,000, respectively.  

\subsubsection{test sets}\label{sec:testsets}
We evaluate the performance of our model independently of the approach we took to generate our datasets. Thus, our test sets (the holdout sample) were composed of two types of instances: \\ \noindent \textbf{Test set 1:} 
a $36,000$ size set based on our sampling technique in Section~\ref{sec:data}. This test set is valuable due to its complexity of inter-arrival and service time distribution (e.g., a wide range of SCV). For this test set, we allow mixing small and large  
inter-arrival and service time SCV values within a cycle. This data set's range of SCV values is $[0.004,394.2]$; for more details regarding the SCV values, see Appendix~\ref{append:SCV}. 

The arrival rates are sampled according to our sampling technique described in Section~\ref{sec:arrival_pattern_sampling}. Our sampling technique also draws the cycle lengths (see Appendix~\ref{append:SCV} for cycle length histogram).  To demonstrate the accuracy of our results, we report them for each  $\bar{\rho}$ interval in the five ranges: $[0.5,0.6]$, $[0.6,0.7]$, $[0.7,0.8]$, $[0.8,0.9]$, $[0.9,1.0]$; each range included $7,200$ test samples.

\noindent  \textbf{Test set 2:} Here also, the arrival rates are sampled according to our sampling technique described in Section~\ref{sec:arrival_pattern_sampling}.  
However, the inter-arrival and service time distributions are not sampled as in the training and validation set. For this purpose, inspired by ~\cite{You19}, we control for the SCV of the inter-arrival and service time distributions as follows. 

We let the inter-arrival and service time follow Log-Normal, Gamma, and Exponential distributions. We set the SCV to 0.25 or 4 for Log-Normal and Gamma, and the exponential SCV is 1. For both inter-arrival and service time distributions, we consider the following cases:  $LN(0.25)$,
$LN(4)$, $G(0.25)$, $G(4)$ and $M$ (see Appendix~\ref{append:dists_notation} for the definition of the distributions).  We consider six cycle utilization levels for each combination of the five inter-arrival and five service-time distributions: $\bar{\rho} \in \{0.5,0.6,0.7,0.8,0.9, 1\}$. In total, we have 150 such combinations. For each combination, we run ten trials, which differ by their arrival rate pattern. Thus, the total size of test set 2 is 1500 samples. Similar to test set 1, our sampling technique also draws the cycle lengths from  Section~\ref{sec:arrival_pattern_sampling} (see Appendix~\ref{append:SCV} for cycle length histogram).
\subsubsection{Evaluation Metrics.}

We use three evaluation metrics for fine-tuning the model and for accuracy evaluation, first presented in~\cite{Sherzer23}. We next present these metrics and note that only the first is used for hyper-parameter selection.

As before, for instance, $i$ in the test data set, let $Y_i$ and $\hat{Y}_i$ represent vectors of size $(l)$ of the true and predicted values of the stationary number of customers in the system, respectively.

Our first metric is the Sum of Absolute Errors (SAE) between the predicted value and the labels:
\begin{equation}  \label{eq:sae}
     SAE_{i,j} =  \sum_{k=0}^{l}|Y_{i,j,k}-\hat{Y}_{i,j,k}|,
\end{equation}

Where $i$ is the $i^{th}$ sample of the set and $j$ refers to the time unit. The average SAE of the size $N$ test set is in the $j^{th}$ time period:
  $   \overline{SAE_j} = \sum_{i=1}^N SAE_{i,j}/N$ ,and the overall average SAE is  $  \overline{SAE} = \sum_{i=1}^N\sum_{j=1}^T SAE_{i,j}/(TN)$.

The $SAE$ is equivalent to Wasserstein-1 measure and to the first term of the loss function in Equation~\eqref{eq:loss}. The advantage of this metric is that (i) it is highly sensitive to any difference between actual and predicted values, and (ii) it produces a single value, which simplifies the comparison between the models, making it particularly suitable for tuning the hyper-parameters of the model.  However, since this metric produces an absolute and not a relative value, it is more difficult to interpret when evaluating model accuracy. 

The second metric, dubbed $PARE$ (Percentile Absolute Relative Error),  measures the average relative error for a specified percentile as follows:   

\begin{align}\label{eq:InvCdf}
PARE_{i, j}(Y,\hat{Y}, percentile) = 100 \frac{|F_{Y_{i,j}}^{-1}(percentile)-F_{\hat{{Y_{i,j}}}}^{-1}(percentile)|}{F_{Y_{i,j}}^{-1}(percentile)},
\end{align}

Where  $F(\cdot)$ is the CDF of $Y_{i,j}$ of the $i^{th}$ sample and the $j^{th}$ period. The average PARE of the size $N$ test set is in the $j^{th}$ time period:
  $   \overline{PARE_j} = \sum_{i=1}^N PARE_{i,j}/N$ ,and the overall average PARE is  $  \overline{PARE} = \sum_{i=1}^N\sum_{j=1}^T PARE_{i,j}/(TN)$.

This metric is effective in detecting differences at any specified percentile, including the tails of the two distributions. As noted earlier, accurate estimates of tail probabilities are important in many queueing applications since they directly relate to system service level measures. We therefore use the following six percentile values in our evaluations: $25\%, 50\%, 75\%,90\%, 99\%, 99.9\%$.   

Our third metric is the $\%$ Relative Error of the Mean (REM) number of customers in the system. The REM of the $i^{th}$ sample and $j^{th}$ time period is:

\begin{align} \label{eq:REM}
REM_{i,j} = 100\frac{|\sum_{k=0}^{l}k(Y_{i,j,k}-\hat{Y}_{i,j,k})|}{\sum_{j=0}^{l-1}k\hat{Y}_{i,j,k}}.
\end{align}

The average REM of the size $N$ test set is in the $j^{th}$ time period:
  $   \overline{REM_j} = \sum_{i=1}^N REM_{i,j}/N$ ,and the overall average REM is  $  \overline{REM} = \sum_{i=1}^N\sum_{j=1}^T REM_{i,j}/(TN)$.

The advantage of this measure is that it summarizes the model accuracy for the most popular performance measure for queueing systems - the average number of customers in the queue. 

\subsubsection{Model Training and Tuning.}\label{sec:Model_tunning}\label{sec:tunnig}

We use the following sequence of steps to train our model. Once both training and validation datasets are sampled and pre-processed, including the output computation, we seek the best-performing hyperparams of the model based on the evaluation of the validation set. First and foremost, the values of $n_{arrival}$ and $n_{service}$. We initially use only the first moments and then increase the values of $n_{arrival}$ and $n_{service}$ together by one each time until as long as the performance of our model keeps increasing. The results of the moment analysis are presented in Section~\ref{sec:mom_analysis}.  For each ($n_{arrival}$, $n_{service}$) combination, we fine-tune the hyper-parameters of the model, i.e., choose the best hyper-parameters (e.g., learning-rate schedule, batch size, etc.) based on the validation set. As mentioned above, different hyper-parameter settings were compared via the $\overline{SAE}$ (see Equation~\eqref{eq:sae}). For a full description of the hyper-parameter fine-tuning process, see Appendix~\ref{append:Fine-tuning}. 

Our model was trained on an NVidia A100 GPU. We used an RNN model with an LSTM. The RNN architecture comprises four hidden layers of 128 nodes. We used the Adam optimizer to update the network weights based on the training data with weight decay as a regularization factor\footnote{Weight decay is a regularization technique that adds a small penalty, typically the L2 norm of the weights (all weights), to the loss function.} to avoid over-fitting (for more details about the Adam optimizer see ~\cite{8624183} and references therein).  

The corresponding weight decay value for avoiding over-fitting was $10^{-5}$, 
while the selected (optimal) training batch size was $B=32$. 
The learning rate was scheduled to decay exponentially, starting from 0.001; the number of trained periods was 60. The training was done in Pytorch in a Python environment.


\subsection{Comparing to diffusion approximation}\label{sec:diffusion}
This section discusses alternative methods that approximate transient models, such as those described in this paper. For this purpose, we compare our performances with a diffusion model proposed by Duda in~\cite{1146391}. (We chose their study as we are unfamiliar with any $G(t)/GI/1$ diffusion process with the same arrival process as ours.) We followed their experiment, in which they reported the accuracy of their results. Experiment 5 provides the exact details. Generally speaking,  the method in~\cite{1146391} computes the distribution of the transient number of customers of a $GI/GI/1$  system. As our model is more general (specifically, the arrival process may change over time), our NN model can also make inferences over the $GI/GI/1$  system. By choosing m=1 periods, we have a single renewal arrival process over T, as in the $GI/GI/1$  system. This allows a natural and precise comparison between our method and diffusion approximations.  

We also provide the fluid accuracy as a reference point within the first five experiments. In the fluid model, each customer is considered as a tiny drop, where each
drop is served in zero time.  These models are commonly used in transient queueing systems (see~\cite{8624183, Chen1997, SHERZER201716}).

\subsection{Number of training periods vs. number of inference periods}\label{sec:inference_epcoh}

In this study, the number of training periods is 60. However, this does not mean the model is restricted to 60 periods for inference. We next present a method for making predictions beyond 60 training periods.  As mentioned in Section~\ref{sec:network}, provided the right input, the model can make inferences for any number of periods. The number of periods in the input will also be the number of output periods.  However, it is unclear what the accuracy will be in periods beyond the training domain. For this purpose, we also conduct a performance evaluation of our accuracy beyond the training periods.

\begin{remark}
      A different approach to deal with this prediction beyond the 60 trained periods is to take the output of the $60^{th}$ period as the initial distribution input for the next 60 periods (i.e., periods 61-120). This method can be repeated recursively \textit{ad infinitum}. We expect accuracy to decay in this method, as the age of inter-arrival and service time distributions are assumed to be 0 whenever we make inferences for the next 60 periods by design. However, there is no reason for the age of inter-arrival and service time distributions to be exactly 0. Empirically, this method is somewhat less accurate than the one above. 
\end{remark}

\subsection{Experiment design details}\label{sec:segmentation} 

When evaluating the \textit{MBRNN} accuracy, we examine values of $\bar{\rho}$ and different SCV  (i.e., $V(\cdot)/E(\cdot)^2$) for both the inter-arrival and service time distributions (see~\cite{https://doi.org/10.1111/j.1937-5956.1993.tb00094.x, https://doi.org/10.1002/nav.22010}). The main idea is that both SCV and $\bar{\rho}$ are key factors in the queueing analysis. Thus, we wish to show that our \textit{MBRNN} performs well for a broad domain for both. Since in test set 1, we have a mixture of inter-arrival distribution within a single sample, we also have multiple SCV values. As such, the effect of SCV values on the accuracy will occur only via test set 2. We also compare our accuracy with this of other methods as discussed in Section~\ref{sec:diffusion} and examine the accuracy beyond the training periods as proposed in Section~\ref{sec:inference_epcoh}.
We, therefore conduct the following experiments:

\begin{itemize}
    \item \textbf{Experiment 1:} Overall accuracy of test set 1.  We compute $\overline{SAE}$, $\overline{PARE}$ for all six percentiles, and $\overline{REM}$  in test set 1. We take the average value at each of the five $\bar{\rho}$ ranges described in Section~\ref{sec:testsets}.
    \item \textbf{Experiment 2:} Accuracy as a function of time of test set 1.  We compute $\overline{SAE_j}$, $\overline{PARE_j}$ for all six percentiles, and $\overline{REM_j}$ as a function of the time period $j$, where $1 \leq j \leq 60$ for test set 1 at each of the $\bar{\rho}$ ranges described in Section~\ref{sec:testsets}.
    \item  \textbf{Experiment 3:} Overall accuracy of test set 2. We compute $\overline{SAE}$, $\overline{PARE}$ for all six percentiles, and $\overline{REM}$ in test set 2 at each of the $\bar{\rho}$  and SCV  ranges as described in Section~\ref{sec:testsets}.
    \item \textbf{Experiment 4:} Accuracy as a function of time of test set 2. We compute $\overline{SAE_j}$, $\overline{PARE_j}$, and $\overline{REM_j}$ as a function of the time period $j$, where $1 \leq j \leq 60$ for test set 2 at each of the $\bar{\rho}$  and SCV ranges as described in Section~\ref{sec:testsets}. (We do not present the $\overline{PARE_j}$ values for different SCV values, as they mount up to many combinations of percentiles, time, and SCV values, and results in a very exhaustive analysis.) 
    \item \textbf{Experiment 5:} Compering to other methods.  Duda~\cite{1146391} conducted three experiments depicted in Table~\ref{tab:diffusion}. Three $GI/GI/1$ systems are described, which vary by the CSV and the inter-arrival and service time distributions. The utilization level is constant with time and is stated in the columns $\rho$. The first row in Table~\ref{tab:diffusion} is simply the $M(t)/M(t)/1$ system, and hence, the ground truth is exact. They report results on the system's distribution and the expected number of customers. We compare $\overline{SAE}$ and $\overline{REM}$ values. We note~\cite{1146391} reports only the probability of having up to 6 customers in the systems, which does not always sum to 1. As such, the $\overline{SAE}$  value is truncated. Therefore, we do not compute the $\overline{PARE}$ values. 
    \item \textbf{Experiment 6:} Accuracy beyond 60 periods.  We produce an extra test set similar to test set 1, with a size of 1500, over 180 periods. The goal is to test the performance of the untrained periods. We examine the $\overline{SAE_j}$  value as a function of time. Also, we compare the $\overline{SAE_j}$ values of the first, second, and last 60 periods.   
    \end{itemize}

\begin{table}[htbp]
  \centering
  \caption{Diffusion experiments}
    \begin{tabular}{|c|c|c|c|c|c|c|}
    \toprule
          &       & \multicolumn{2}{c|}{Inter arrival } & \multicolumn{2}{c|}{Service} &  \\
    \midrule
    \#    & Initial state & Distribution  & SCV   & Distribution  & SCV   & $\bar{\rho}$ \\
    \midrule
    1     & 5     & Exponential & 1.0  & Exponential & 1.0  & 0.5 \\
    \midrule
    2     & 5     & Hyper-Exponential & 0.5  & Hyper-Exponential & 0.5  & 0.5 \\
    \midrule
    3     & 5     & Erlang & 5     & Erlang & 5     & 0.5 \\
    \bottomrule
    \end{tabular}%
  \label{tab:diffusion}%
\end{table}%

\section{Accuracy of the MBRNN Approach}\label{sec:result}

In this section, we present the results of the moment analysis and the results of all six accuracy experiments.

\subsection{Moment analysis}\label{sec:mom_analysis}

\begin{figure}
\centering
\includegraphics[scale=0.42]{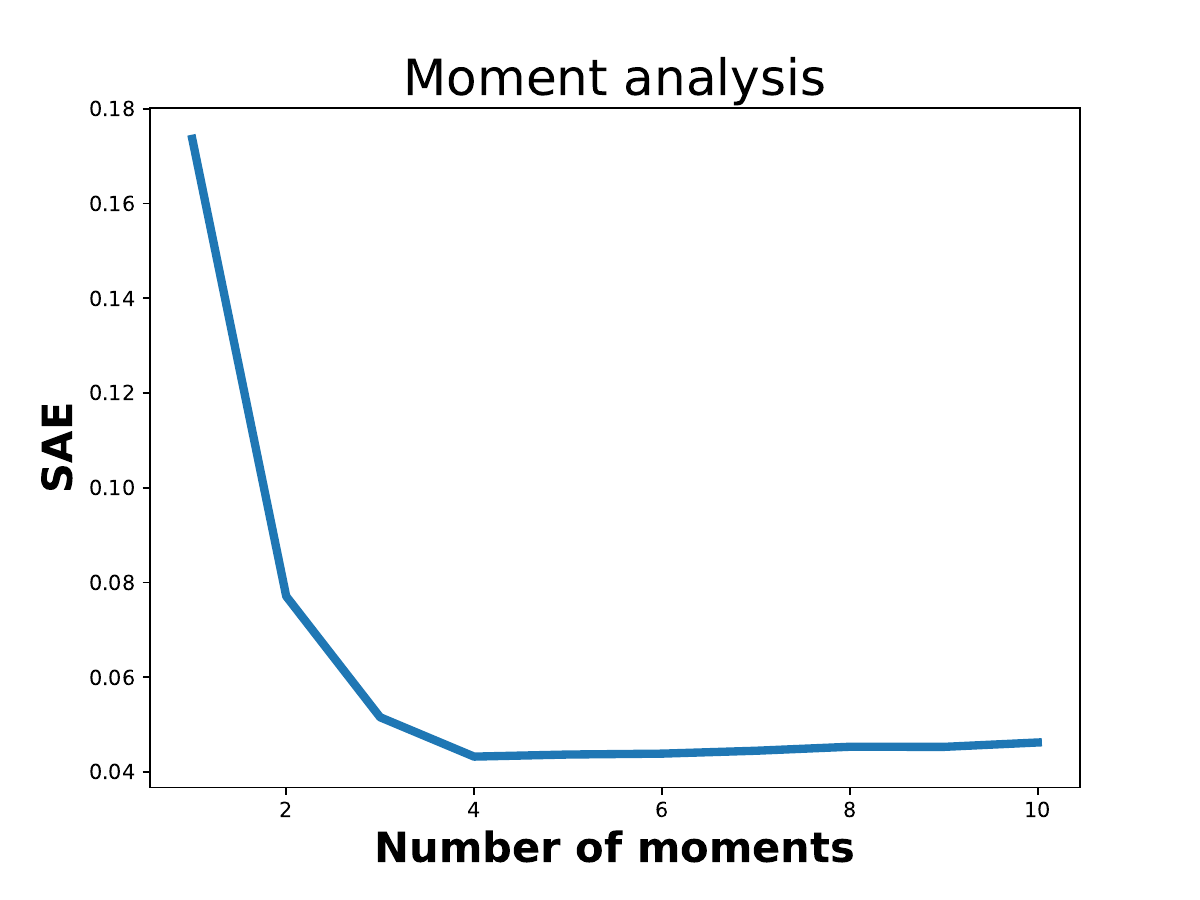}
\caption{Computing the $\overline{SAE}$  as a function of the number of inter-arrival and service time moments.}
\label{fig:mom_analysis}
\end{figure}

As discussed in Section~\ref{sec:Model_tunning}, we examine the effect of the number of inter-arrival and service time moments inserted into the model on the overall accuracy with respect to test set 1. The resulting SAE are depicted in Figure~\ref{fig:mom_analysis}. The result suggests that the optimal point is four moments with an $\overline{SAE}$ of 0.043\footnote{As such, similar results are given by the other two metrics as well as they are correlated.}. This does not mean that the fifth moment and beyond do not affect queue dynamics. However, within the model's constraints, which rely on labeling via simulations and not exact analysis, and an ML approach, which also inflicts some error, we cannot extract relevant information from the $5^{th}$ and beyond moments. 
In~\cite{Sherzer23}, where a similar experiment took place, the system benefited from the fifth moment in the stationary $GI/GI/1$ system. There can be many reasons for that. For instance, this may be due to the difference between the stationary and transient systems. However, we believe this is merely due to the labeling accuracy, where in~\cite{Sherzer23}, labeling was done numerically with greater accuracy. As such, the model is less sensitive to labeling errors and can extract value from the fifth moment.

\subsection{Experiment 1: Overall accuracy of test set 1}\label{sec:exp1}

Table~\ref{tab:sae1_rem1_rho} depicts the $\overline{SAE}$ and $\overline{REM}$ results, where the $\pm$ indicates the 95\%  half confidence interval length. Surprisingly, the $\overline{SAE}$ and $\overline{REM}$  results are in opposite directions. The $\overline{SAE}$ accuracy decreases while the $\overline{REM}$ accuracy increases with $\bar{\rho}$. Nevertheless, both demonstrate solid performance, as the largest REM value is smaller than 3\%. Furthermore, as expected, the \textit{MBRNN} model performs much better than the fluid approximation.

\begin{figure}
\centering
\includegraphics[scale=0.42]{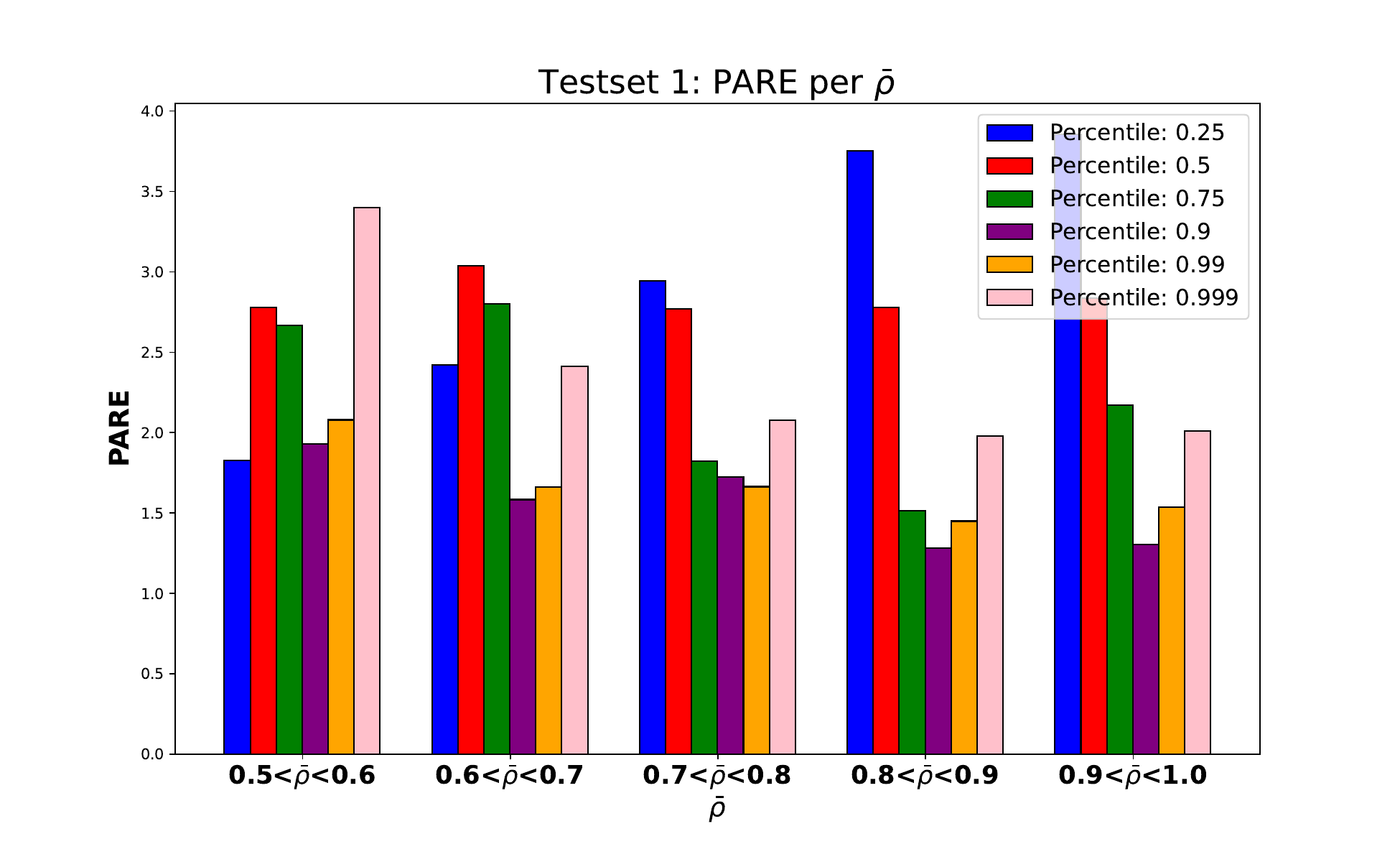}
\caption{PARE by percentile and $\bar{\rho}$.}
\label{fig:PARE1}
\end{figure}

The PARE results are depicted in Figure~\ref{fig:PARE1}. The largest error, for  percentile $25^{th}$ and $0.9 \leq \bar{\rho} <1$ is smaller than 4\%. Surprisinigly, we observe that the $25^{th}$ percentile $\overline{PARE}$ value is increasing with $\bar{\rho}$, while the $99.9^{th}$  percentile $\overline{PARE}$ value is decreasing. 

\begin{table}[htbp]
  \centering
  \caption{Test set 1: accuracy results. }
    \begin{tabular}{|r|r|r|r|r|r|}
    \toprule
          & \multicolumn{2}{c|}{$\bar{\rho}$} &       & \multicolumn{2}{c|}{$\overline{REM}$} \\
    \midrule
    \multicolumn{1}{|l|}{\#} & \multicolumn{1}{l|}{LB} & \multicolumn{1}{l|}{UB} & \multicolumn{1}{l|}{$\overline{SAE}$} & \multicolumn{1}{l|}{\textit{MBRNN}} & \multicolumn{1}{l|}{Fluid} \\
    \midrule
    1     & 0.5   & 0.6   & 0.044$\pm$ 0.006 & 2.71 $\pm$ 1.62 & 35.44  $\pm$ 27.11\\
    \midrule
    2     & 0.6   & 0.7   & 0.045 $\pm$ 0.005 & 2.26 $\pm$ 1.24 & 29.38 $\pm$ 20.52\\
    \midrule
    3     & 0.7   & 0.8   & 0.046 $\pm$ 0.006  & 2.06 $\pm$ 1.26 & 24.72 $\pm$ 15.73\\
    \midrule
    4     & 0.8   & 0.9   & 0.047 $\pm$ 0.006 & 1.92 $\pm$ 1.1 & 22.87 $\pm$ 13.77\\
    \midrule
    5     & 0.9   & 1     & 0.048 $\pm$ 0.005 & 1.82 $\pm$ 0.97 & 24.11 $\pm$ 19.55\\
    \bottomrule
    \end{tabular}%
  \label{tab:sae1_rem1_rho}%
\end{table}%

\subsection{Experiment 2: Accuracy as a function of time of test set 1}\label{sec:exp2}

\begin{figure}[h]
        \centering
        \begin{subfigure}[b]{0.4775\textwidth}
            \centering
        \includegraphics[width=7.5cm,height=4.592cm]{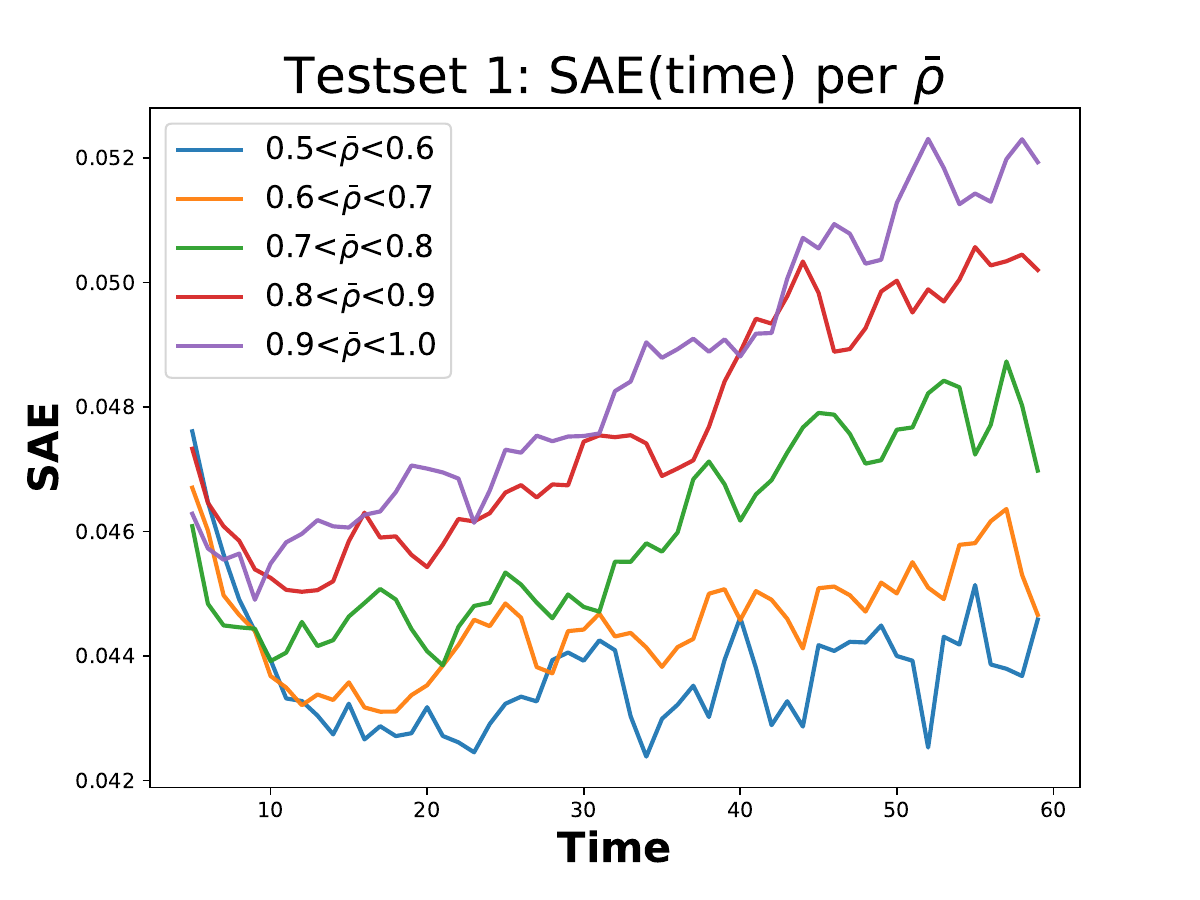}
            \caption[]%
            {{\small SAE}}    
            \label{fig:SAE_time_test1_rho}
        \end{subfigure}
        \hfill
        \begin{subfigure}[b]{0.4375\textwidth}  
            \centering 
            \includegraphics[width=7.5cm,height=4.592cm]{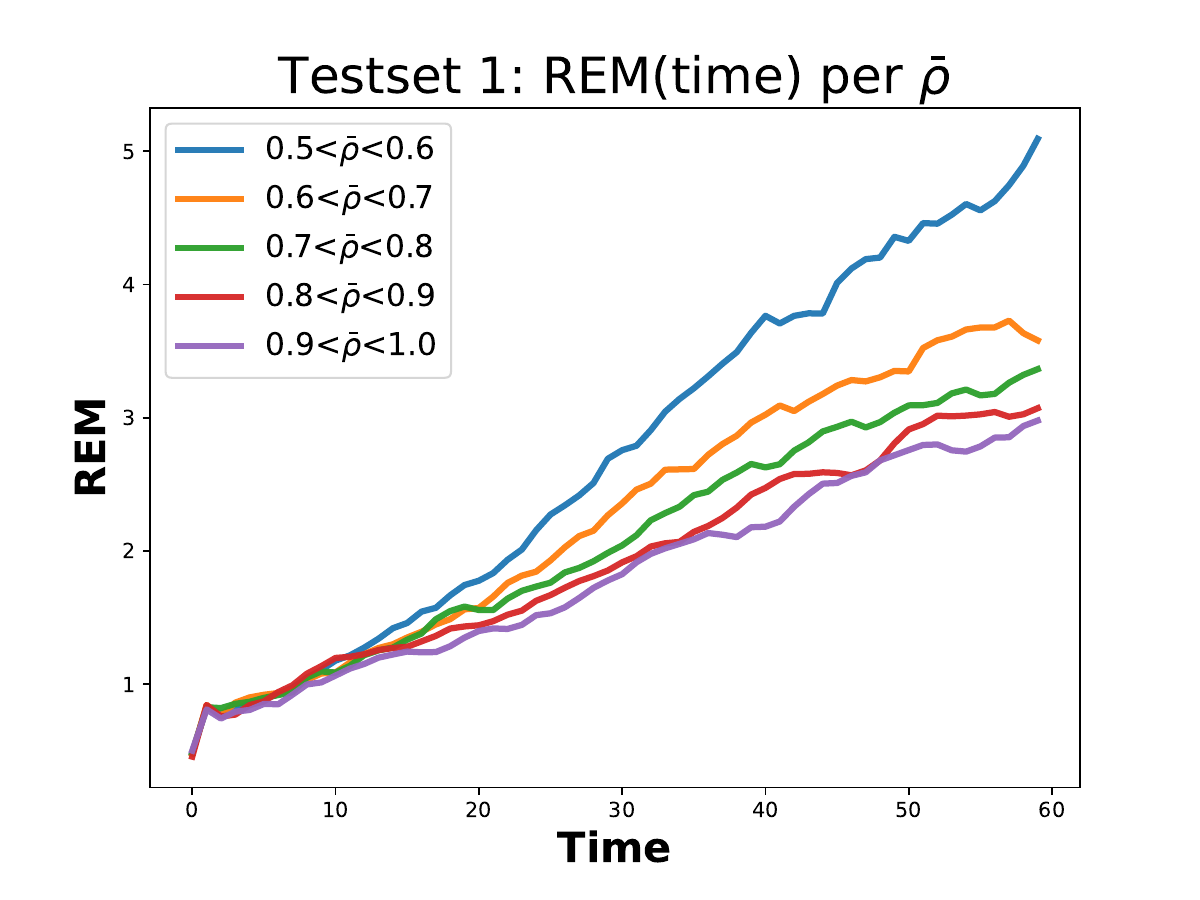}
            \caption[]%
            {{\small REM}}    
            \label{fig:REM_time_test1_rho}
        \end{subfigure}
        \caption{Test set 1: accuracy results as a function of time}%
    \label{fig:test1_SAE_REM}
    \end{figure}

\begin{figure}
\centering
\includegraphics[scale=0.42]{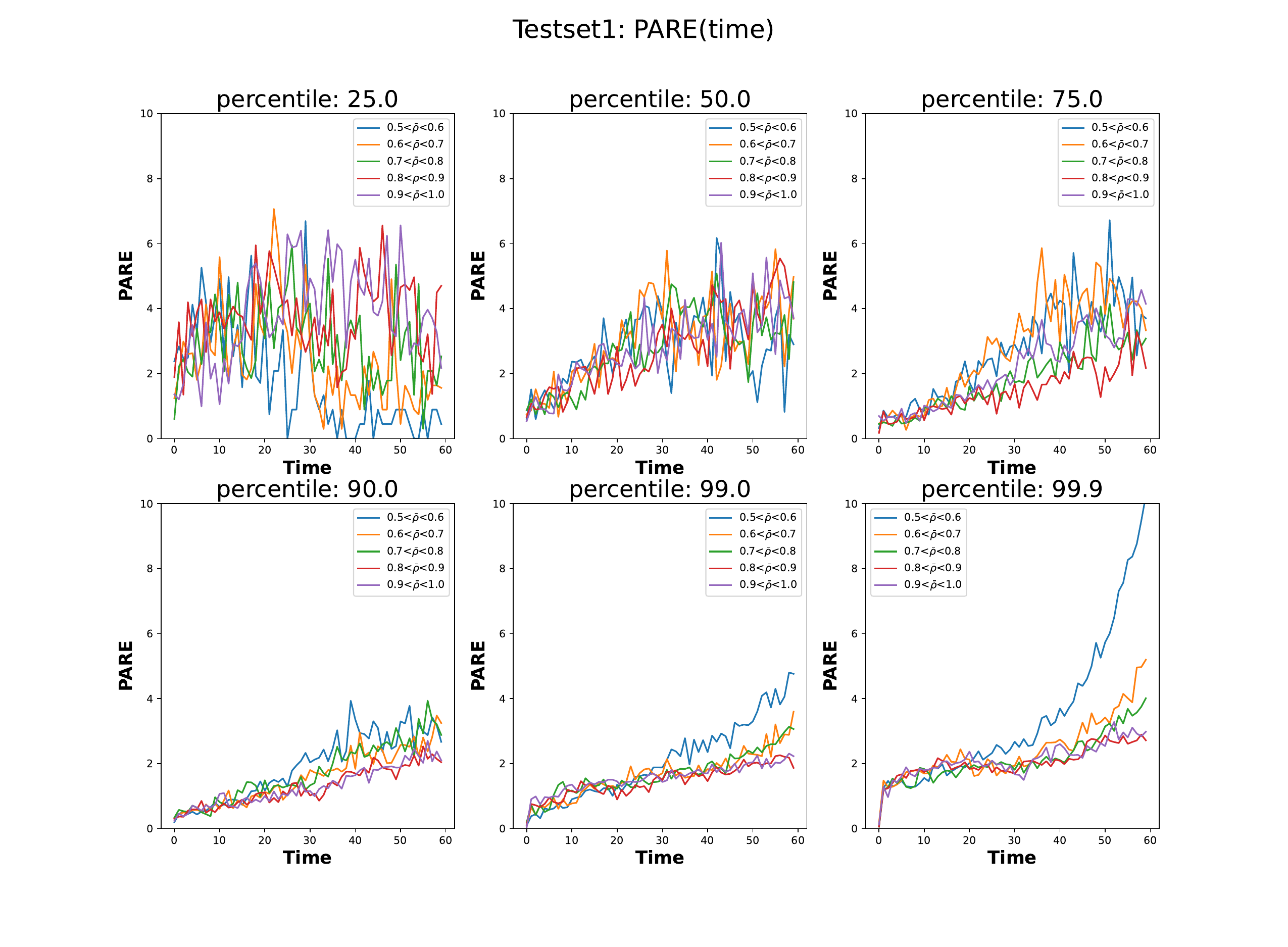}
\caption{PARE by percentile and $\bar{\rho}$ as a function of time.}
\label{fig:PARE1_time}
\end{figure}

Figures~\ref{fig:SAE_time_test1_rho}, \ref{fig:REM_time_test1_rho} and~\ref{fig:PARE1_time}, represent the $\overline{SAE_j}$, $\overline{REM_j}$, and $\overline{PARE_j}$ errors as a function of the time period $j\leq 60$, respectively. All performance measures demonstrate a certain decay of accuracy with time. Similar to the results in Experiment 1,  $\overline{SAE_j}$ performs better for low average cycle utilization levels, while the opposite is true for the $\overline{REM_j}$. For $\overline{PARE_j}$, we observe a steep increase for percentiles  (i.e., 99 and 99.9) for $t>40$. We do not observe a monotone behavior of the $\overline{PARE_j}$  and percentile values. Finally, while our labeling error is decreasing with time (see Figure~\ref{fig:CI}) the MBRNN error is increasing. This may be due to the fact that each period relies on the prediction of previous results. Hence, the hidden input accumulates with greater errors with time.

\subsection{Experiment 3: Overall accuracy of test set 2}\label{sec:exp3}

Table~\ref{tab:sae1_rem2_rho} depicts the $\overline{SAE}$ and $\overline{REM}$ results for test set 2. These are slightly better than the results in Table~\ref{tab:sae1_rem1_rho} for test set 1. These results indicate the robustness of our method, as the underlying inter-arrival and service time distributions for test set 2 were generated differently than our training set, yet no decrease in the accuracy occurred. 

\begin{table}[htbp]
  \centering
  \caption{Test set 2: accuracy results by $\bar{\rho}$}
    \begin{tabular}{|r|r|r|r|r|}
    \toprule
          &       &       & \multicolumn{2}{c|}{$\overline{REM}$} \\
    \midrule
    \multicolumn{1}{|l|}{\#} & \multicolumn{1}{l|}{$\bar{\rho}$ } & \multicolumn{1}{l|}{$\overline{SAE}$} & \multicolumn{1}{l|}{\textit{MBRNN}} & \multicolumn{1}{l|}{Fluid} \\
    \midrule
    1     & 0.5   & 0.037 $\pm$ 0.002 & 1.77 $\pm$ 1.4 & 34.60 $\pm$ 32.15\\
    \midrule
    2     & 0.6   & 0.038 $\pm$ 0.001 & 1.40 $\pm$ 0.9 & 28.09 $\pm$ 24.16\\
    \midrule
    3     & 0.7   & 0.040 $\pm$ 0.001 & 1.27 $\pm$ 0.9 & 22.03 $\pm$ 16.77\\
    \midrule
    4     & 0.8   & 0.041 $\pm$ 0.001 & 1.18 $\pm$ 0.9 & 19.83 $\pm$ 15.44\\
    \midrule
    5     & 0.9   & 0.042 $\pm$ 0.001 & 1.14 $\pm$ 0.9 & 16.52 $\pm$ 11.07\\
    \midrule
    6     & 1     & 0.044  $\pm$ 0.001 & 1.15 $\pm$ 1.1 & 15.94 $\pm$ 11.05\\
    \bottomrule
    \end{tabular}%
  \label{tab:sae1_rem2_rho}%
\end{table}%

In Table~\ref{tab:sae1_rem2_SCV}, we stratify the results as in Table~\ref{tab:sae1_rem2_rho} by the SCV of both inter-arrival and service time distributions. The results suggest that the least accurate SCV is 4, and the most accurate is 1. This is not surprising, as a large SCV implies a large variance, which is typically more challenging. 

\begin{table}[htbp]
  \centering
  \caption{Test set 2: accuracy results by SCV}
    \begin{tabular}{|c|c|c|c|c|c|c|c|c|c|}
    \toprule
    Experiement          & 1     & 2     & 3     & 4     & 5     & 6     & 7     & 8 & 9 \\
    \midrule
    Inter-arrival SCV  & 0.25  & 0.25  & 0.25  & 4     & 4     & 4     & 1     & 1     & 1 \\
    \midrule
    Service SCV  & 0.25  & 4     & 1     & 0.25  & 4     & 1     & 0.25  & 4     & 1 \\
    \midrule
    $\overline{SAE}$   & 0.037 & 0.042 & 0.035 & 0.042 & 0.047 & 0.042 & 0.040 & 0.042 & 0.039 \\
    \midrule
    $\overline{REM}$: \textit{MBRNN} & 1.22  & 1.43  & 1.34  & 1.55  & 1.79  & 1.81  & 0.53  & 0.64  & 0.48 \\
    \midrule
    $\overline{REM}$: Fluid & 27.04 & 27.75 & 26.59 & 26.45 & 27.02 & 25.64 & 7.40  & 6.16  & 6.37 \\
    \bottomrule
    \end{tabular}%
  \label{tab:sae1_rem2_SCV}%
\end{table}%

\begin{figure}
\centering
\includegraphics[scale=0.42]{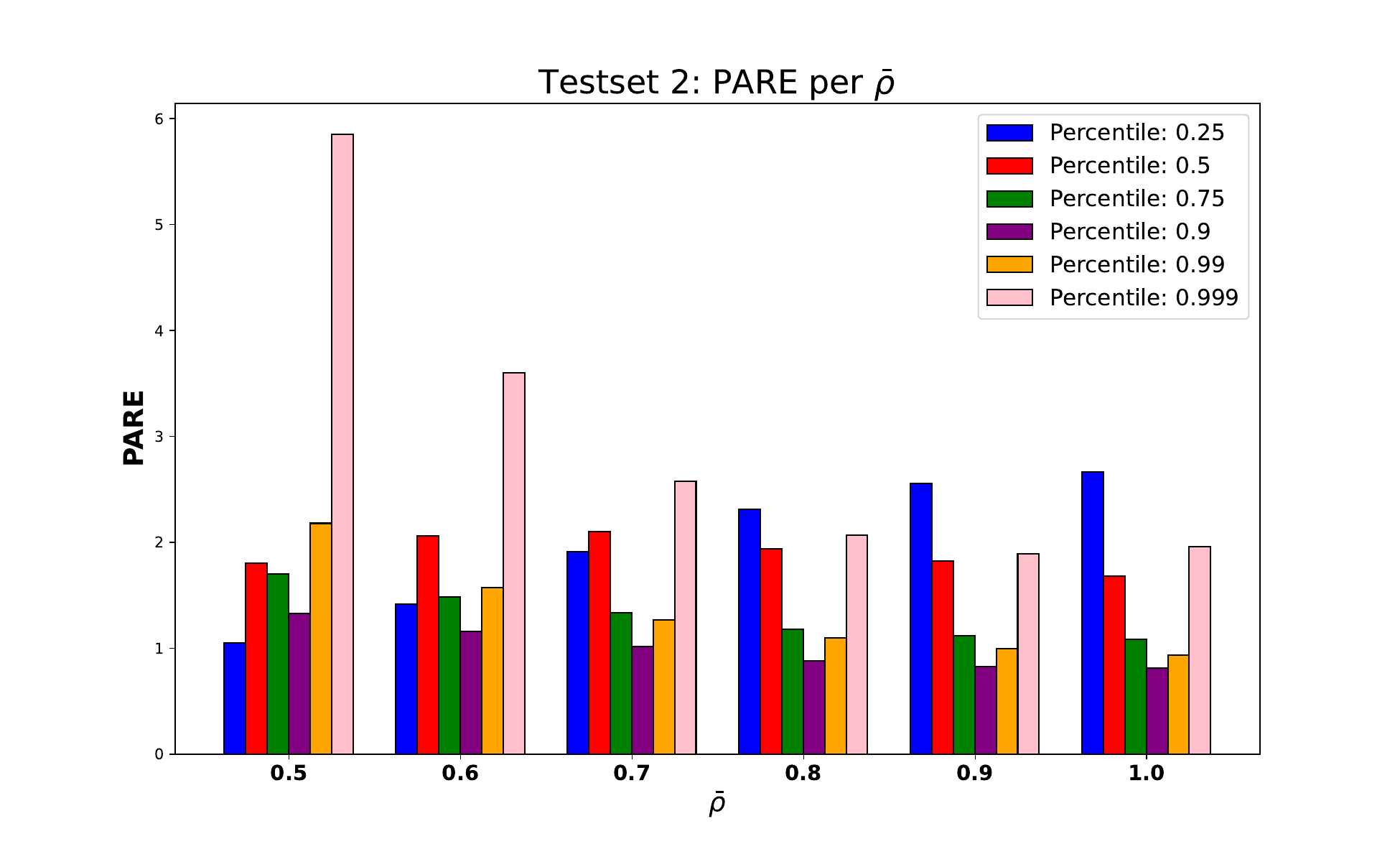}
\caption{PARE by percentile and $\bar{\rho}$.}
\label{fig:PARE2_rho}
\end{figure}
\begin{figure}
\centering
\includegraphics[scale=0.42]{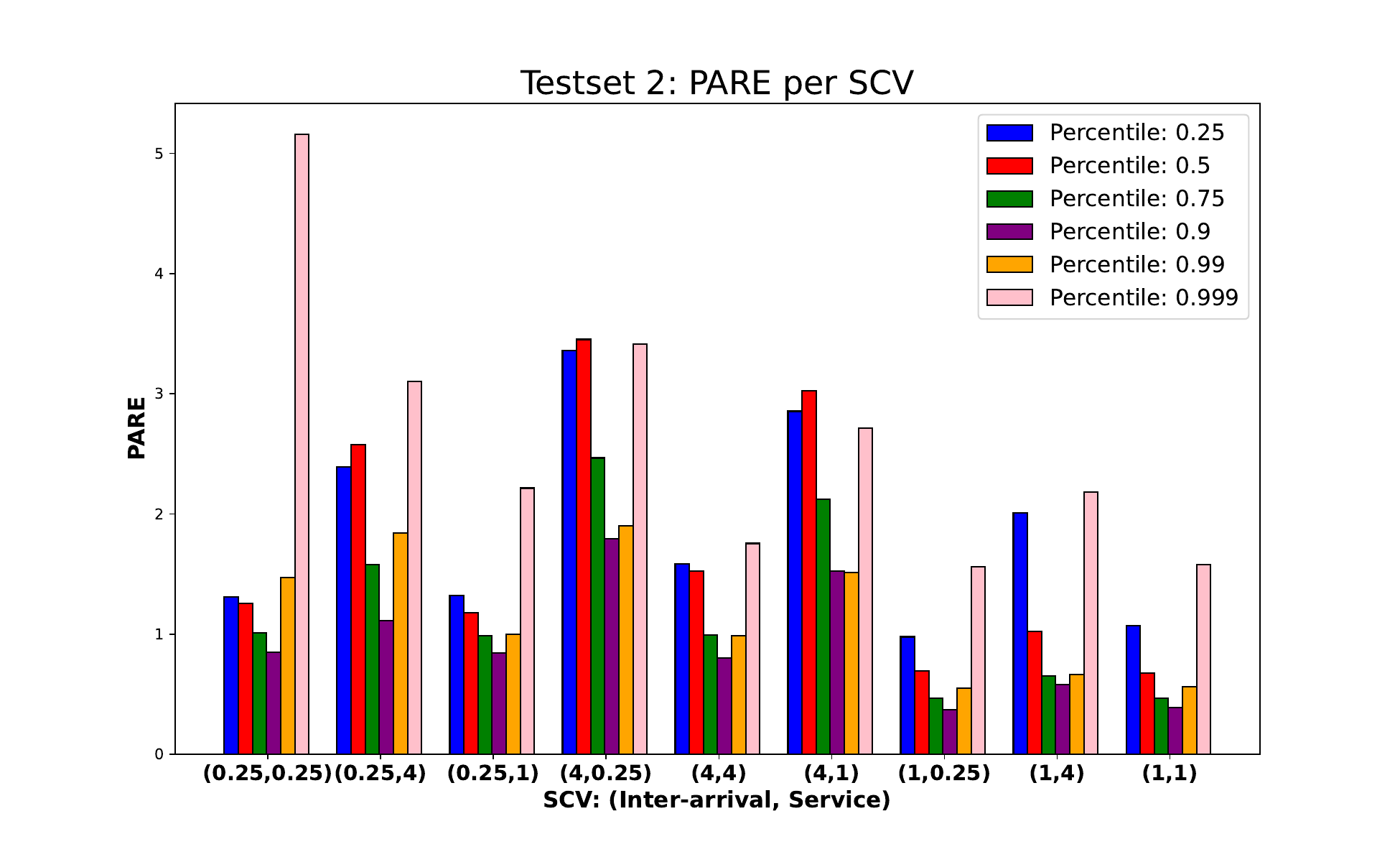}
\caption{PARE by percentile and SCV.}
\label{fig:PARE2_SCV}
\end{figure}

The $\overline{PARE}$ results are depicted in Figures~\ref{fig:PARE2_rho} and~\ref{fig:PARE2_SCV}. In Figure~\ref{fig:PARE2_rho}, we slice the data into different domains of $\bar{\rho}$, and we see similar results as in Figure~\ref{fig:PARE1}, which is the equivalent chart for Test set 1. Figure~\ref{fig:PARE2_SCV}, where we control the SCV of inter-arrival and service time distributions. Similar to results from Table~\ref{tab:sae1_rem2_SCV}, we observe larger $\overline{PARE}$ values where at least one of the distributions SCV is 4. Smaller values of $\overline{PARE}$ occur under the exponential case (i.e., SCV is 1).

\subsection{Experiment 4: Accuracy as a function of time of test set 2}\label{sec:exp4}

\begin{figure}
 \begin{subfigure}{0.49\textwidth}
     \includegraphics[width=\textwidth]{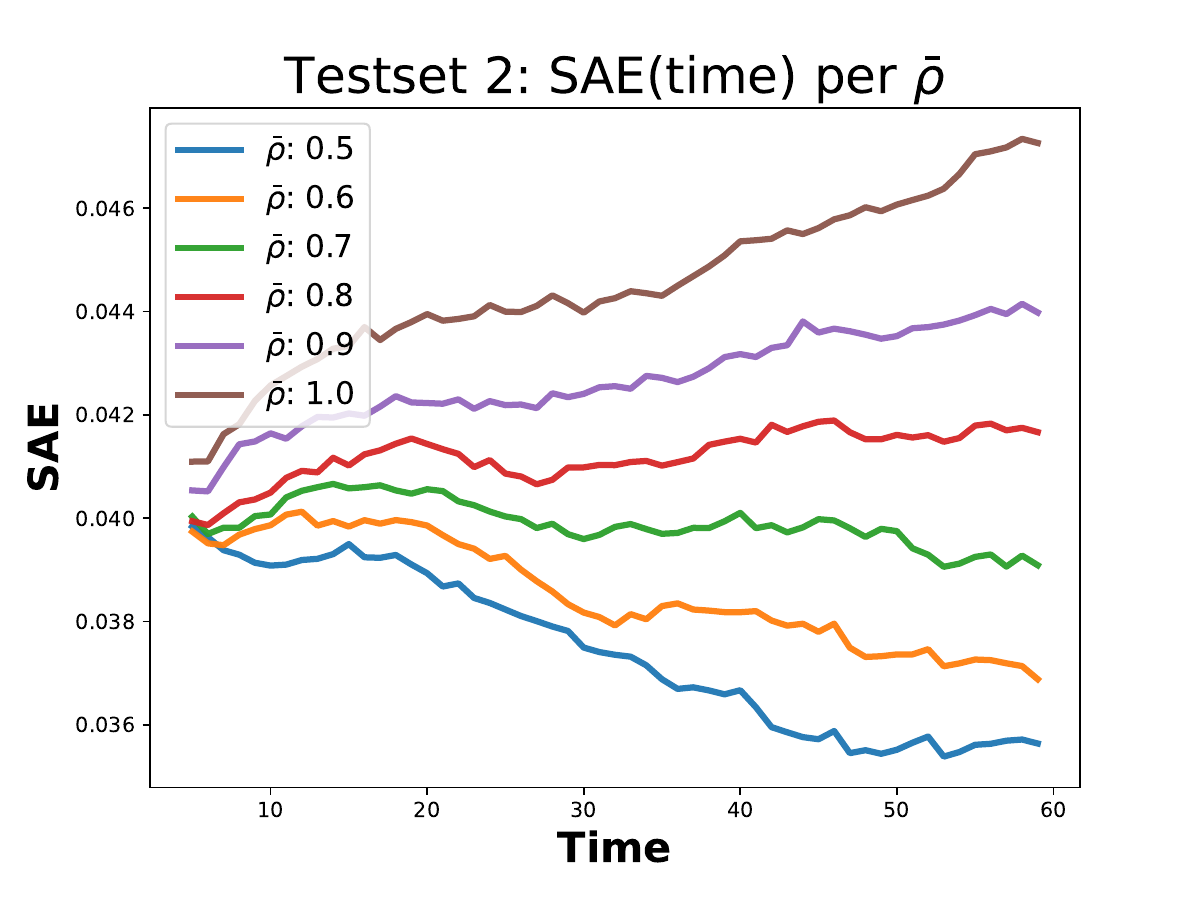}
     \caption{Test set 2: $\overline{SAE}$ per $\bar{\rho}$}
     \label{fig:SAE_time_test2_rho}
 \end{subfigure}
 \hfill
 \begin{subfigure}{0.49\textwidth}
     \includegraphics[width=\textwidth]{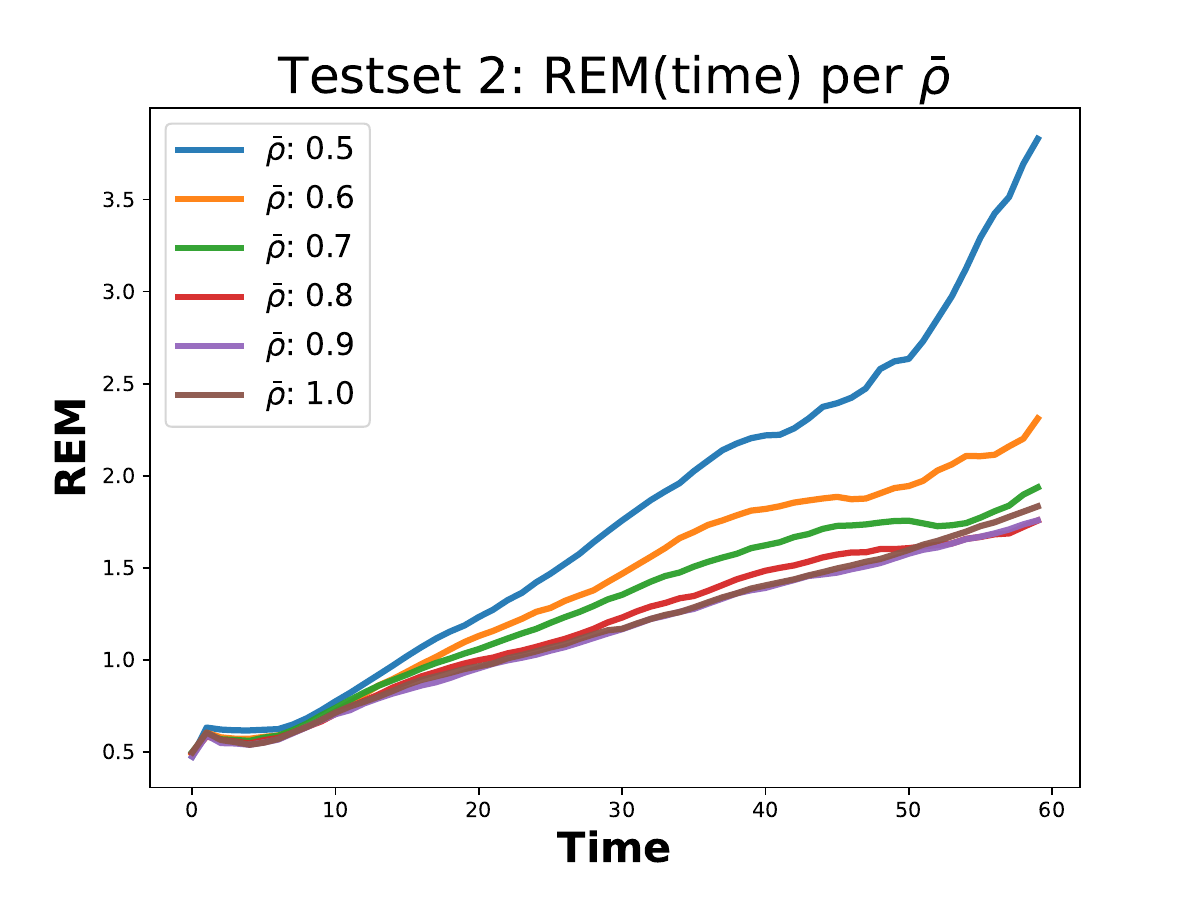}
     \caption{Test set 2: REM per $\bar{\rho}$}
     \label{fig:REM_time_test2_rho}
 \end{subfigure}
 
 \medskip
  \begin{subfigure}{0.49\textwidth}
     \includegraphics[width=\textwidth]{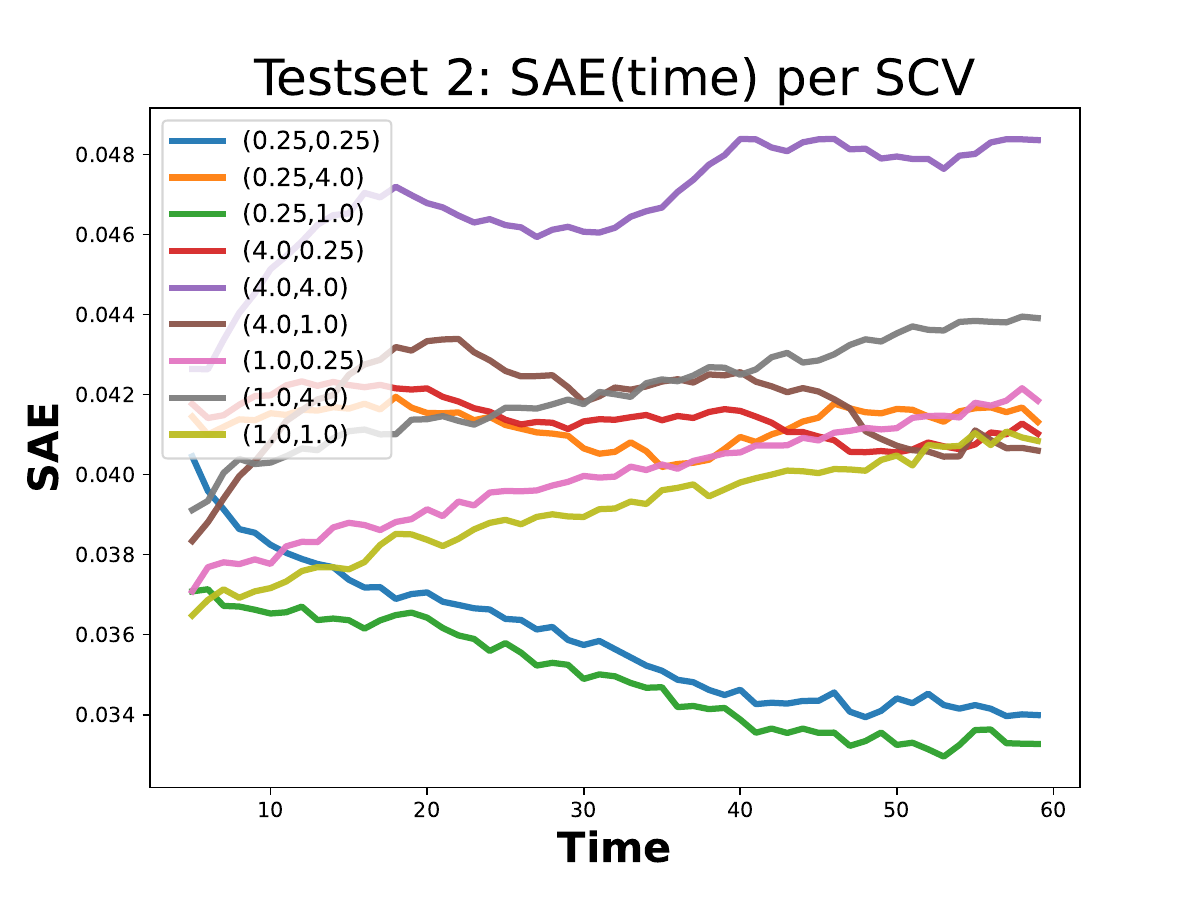}
     \caption{Test set 2: $\overline{SAE}$ per SCV}
     \label{fig:SAE_time_test2_SCV}
 \end{subfigure} 
 \hfill
 \begin{subfigure}{0.49\textwidth}
     \includegraphics[width=\textwidth]{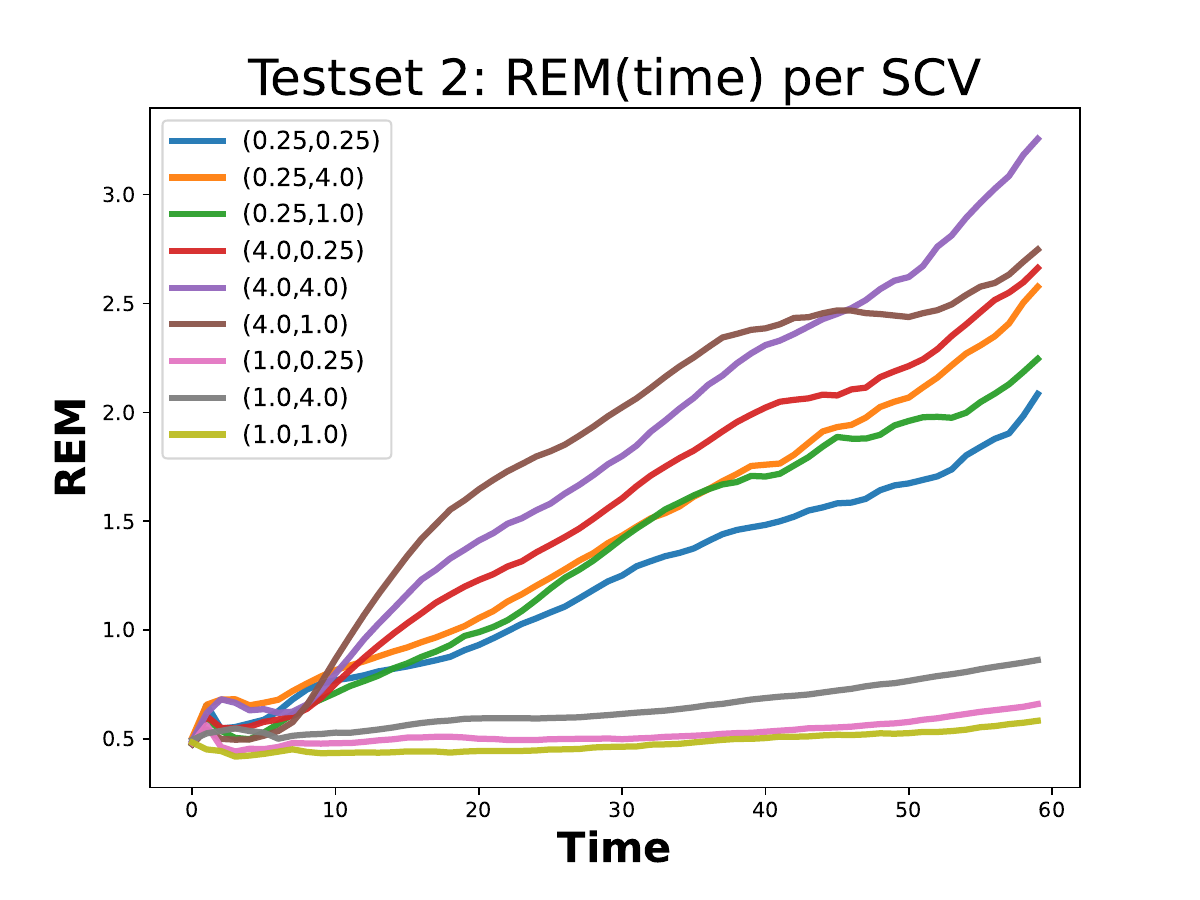}
     \caption{Test set 2: REM per SCV}
     \label{fig:REM_time_test2_SCV}
 \end{subfigure}

 \caption{Test set 2: accuracy results as a function of time}
 \label{Label}

\end{figure}

Figures~\ref{fig:SAE_time_test2_rho} and~\ref{fig:REM_time_test2_rho} depict the $\overline{SAE_j}$ and $\overline{REM_j}$ values as a function of time period $j\leq 60$, respectively. We observe similar behavior as in Figures~\ref{fig:SAE_time_test1_rho} and~\ref{fig:REM_time_test1_rho}. Figures~\ref{fig:SAE_time_test2_SCV} and~\ref{fig:REM_time_test2_SCV} depict the accuracy changes over time and a function of the SCV, support the findings in Experiment 3, which suggest larger accuracy for SCV value of 1, and lower for SCV value of 4.

\begin{figure}
\centering
\includegraphics[scale=0.42]{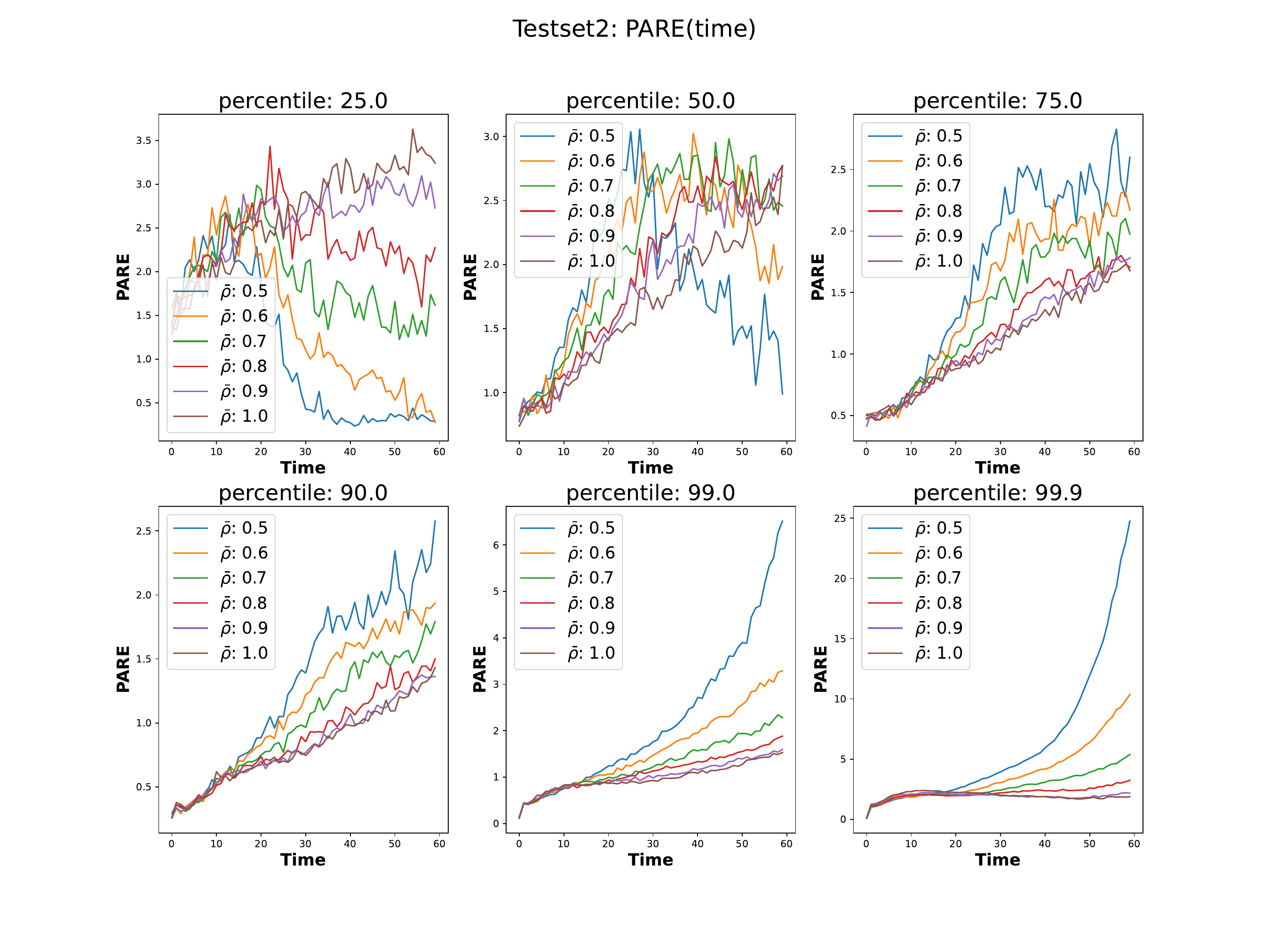}
\caption{$\overline{PARE}$ by percentile and $\bar{\rho}$ as a function of time.}
\label{fig:PARE_time_test_2}
\end{figure}

In Figure~\ref{fig:PARE_time_test_2}, we examine the $\overline{PARE_j}$ values as a function of the time  periods $j\leq 60$. While we observe similar results as in Figure~\ref{fig:PARE1_time}, however, the accuracy decay for the $99.9^{th}$ percentile for $\bar{\rho}=0.5$ is steeper and reaches up to 25\% for $0.5 \leq \rho <0.6$.  These larger errors for the $99^{th}$ percentile are expected for systems with low utilization because then we only have a relatively small sample of these "large deviation" realizations- and thus, our estimation error is larger.

\subsection{Experiment 5: Comparing to other methods}\label{sec:diff_comper}

This experiment examines our accuracy against a diffusion approximation provided in~\cite{1146391}. Table~\ref{tab:diffusion_results} compares the $\overline{SAE}$ and $\overline{REM}$ values. We note that in the first and second rows, where the SCV are 1 and 0.5, respectively, the diffusion performance is very good and is almost comparable to our NN prediction (other than the $\overline{REM}$ for SCV=0.5, where the diffusion approximation is a little better than the NN prediction). However, the accuracy of the diffusion approximation collapses when the SCV is five. In this case, the \textit{MBRNN} prediction outperforms the diffusion approximation by more than $12\%$   in the $\overline{REM}$ measure, and the $\overline{SAE}$ is more than three times smaller. We conclude that the \textit{MBRNN} method outperforms the diffusion approximation in the $GI/GI/1$ system.

\begin{table}[htbp]
  \centering
  \caption{Compering to diffusion approximation}
    \begin{tabular}{|r|r|r|r|r|r|r|}
    \toprule
          & \multicolumn{2}{c|}{SCV} & \multicolumn{2}{c|}{$\overline{SAE}$} & \multicolumn{2}{c|}{$\overline{REM}$} \\
    \midrule
    \multicolumn{1}{|l|}{\#} & \multicolumn{1}{l|}{Inter-arrival} & \multicolumn{1}{l|}{Sevice} & \multicolumn{1}{l|}{Diffusion} & \multicolumn{1}{l|}{\textit{MBRNN}} & \multicolumn{1}{l|}{Diffusion} & \multicolumn{1}{l|}{\textit{MBRNN}} \\
    \midrule
    1     & 1     & 1     & 0.035 & 0.03  & 1.219 & 1.01 \\
    \midrule
    2     & 0.5   & 0.5   & 0.039 & 0.015 & 0.21  & 1.15 \\
    \midrule
    3     & 5     & 5     & 0.286 &  0.08     & 15.594 & 2.71 \\
    \bottomrule
    \end{tabular}%
  \label{tab:diffusion_results}%
\end{table}%

\subsection{Experiment 6: Accuracy beyond 60 periods}

\begin{figure}
\centering
\includegraphics[scale=0.52]{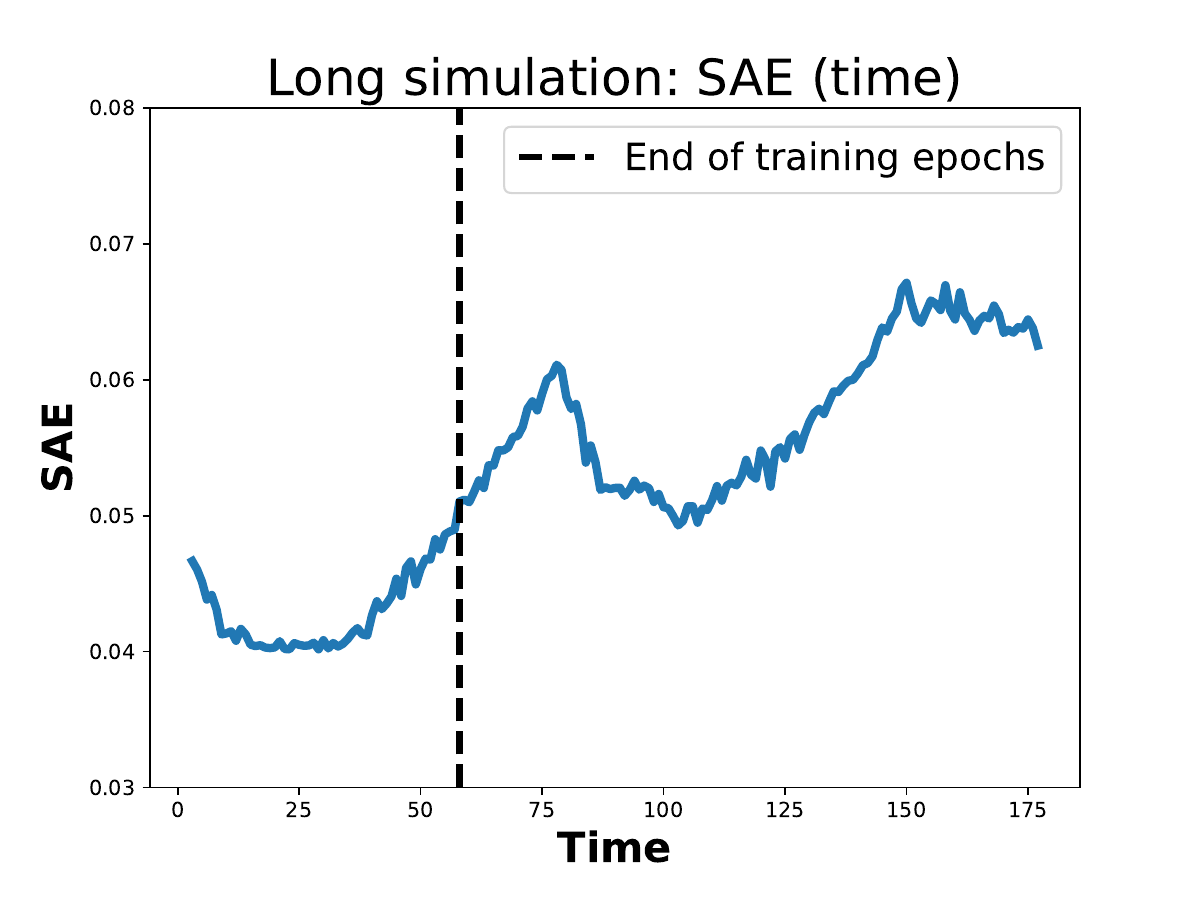}
\caption{SAE time long 180.}
\label{fig:SAE_time_long_180}
\end{figure}

In Figure~\ref{fig:SAE_time_long_180}, we present $\overline{SAE}$ as a function of time up to 180 periods. The results depict only a slight deterioration of the SAE as a function of time.  We also see non-monotonic behavior, probably due to the cyclic nature of the arrival process.

Table~\ref{tab:long_sim_results} depicts the average $\overline{SAE}$, $\overline{PARE}$ and $\overline{REM}$ of the first, second, and last 60 periods. The results showed a mild monotone increase in all metrics. It seems that the decrease in accuracy is not due to the fact that we are beyond the training periods but rather a natural decrease, which also happens within the training periods due to the accumulative error in the hidden input mentioned above. 

\begin{table}[htbp]
  \centering
  \caption{180 periods results}
    \begin{tabular}{|c|c|c|c|c|c|c|c|c|}
    \toprule
          &       & \multicolumn{6}{c|}{$\overline{PARE}$}                     &  \\
    \midrule
    Periods & $\overline{SAE}$   & 25    & 50    & 75    & 90    & 99    & 99.9  & $\overline{REM}$ \\
    \midrule
    1-60  & 0.042 & 2.56  & 2.59  & 2.58  & 2.79  & 2.99  & 3.51  & 2.23 \\
    \midrule
    60-120 & 0.05  & 3.07  & 3.1   & 3.09  & 3.3   & 3.53  & 4.02  & 2.46 \\
    \midrule
    121-180 & 0.055 & 3.68  & 3.71  & 3.74  & 3.91  & 3.94  & 4.43  & 2.67 \\
    \bottomrule
    \end{tabular}%
  \label{tab:long_sim_results}%
\end{table}%

\begin{figure}
\centering
\includegraphics[scale=0.42]{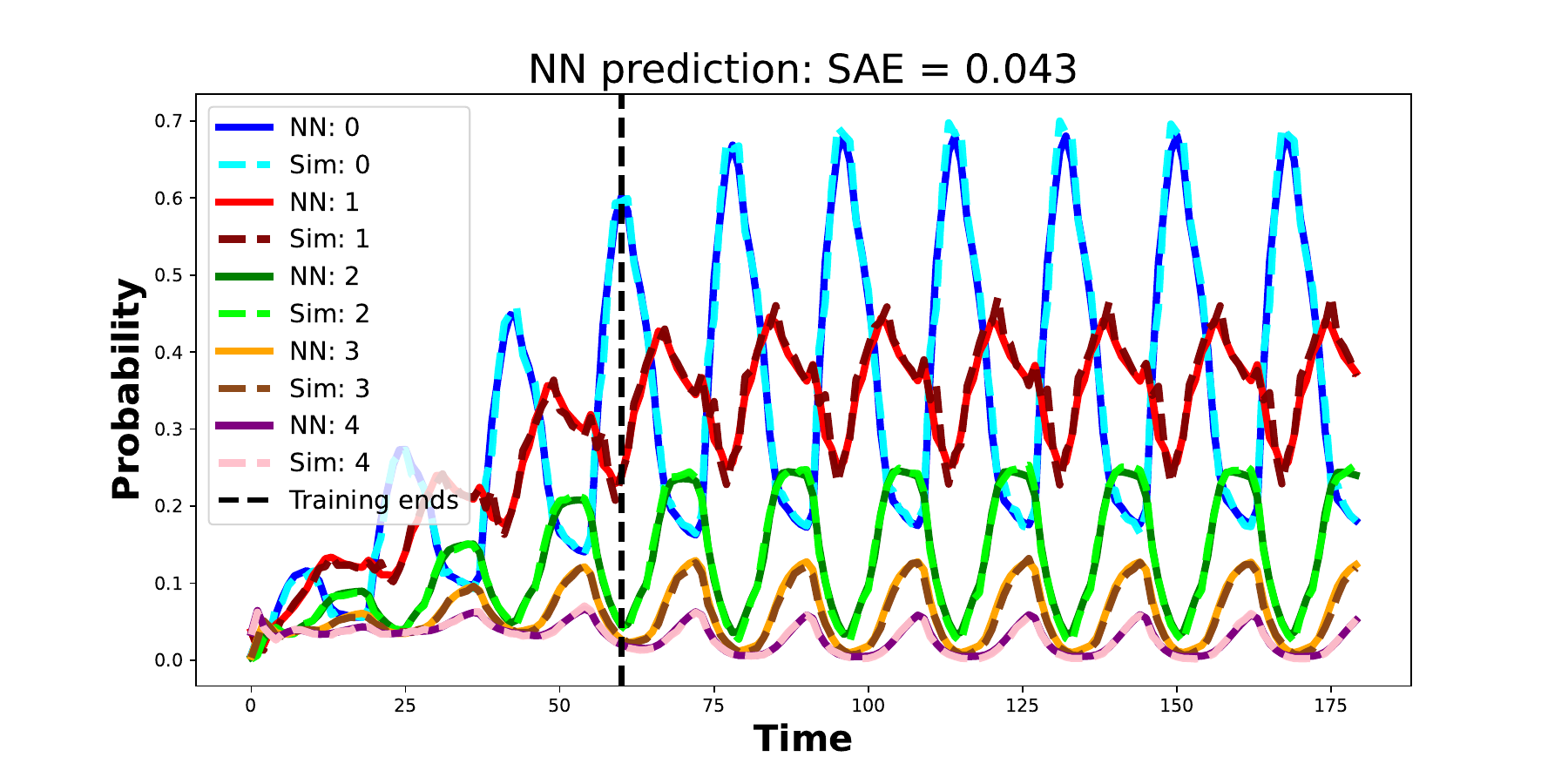}
\caption{long sim example.}
\label{fig:2long_sim_example}
\end{figure}

Finally, for illustration, we present in Figure~\ref{fig:2long_sim_example} an example of the \textit{MBRNN} inference (denoted by "NN" for brevity in the figure). In the example, we plot the probability of having 0 to 4 customers in the system as a function of time for both the \textit{MBRNN} prediction and the simulation. The $\overline{SAE}$ of this example is 0.043 (which is also $\overline{SAE}$ value of test set 1).  As can be seen, the \textit{MBRNN} prediction aligns with the simulation for all values, which serves as the ground truth.

\section{Applications}\label{sec:numeric}
This section aims to demonstrate the effectiveness of the \textit{MBRNN} as a building block in a practical queueing problem. We first  analyze a simple optimization problem and then discuss the data-driven usage of our algorithm for inference.
\subsection{Optimization}\label{subsec:optim}

Consider a $G(t)/GI/1$ system with the arrival process presented in Section~\ref{sec:arrival_proc}. The manager controls the service capacity by adjusting the service rate. There are two types of costs. The first is a linear capacity cost with a cost $C_1$ per unit of capacity. The second is a linear waiting cost $C_2$ per time unit spent in the system per customer. We replace the continuous time problem with a discrete one where we optimize over the number of customers at each $t=1,...,T$. We denote the average number of customers in the system at time $t$ by $e[t]$, which is affected by the service rate. In addition, we require that the probability of having 30 or more customers in the system will not exceed 0.1\% at any $t\in [0,T]$. Thus, the optimization problem we consider is formally:

\begin{equation}
\begin{aligned}\label{eq:constraint}
& \min_{rate}: \quad C_1*rate + C_2*\sum_{t=1}^Te[t] \\
\textrm{s.t.} \quad & \sum_{i \geq 30}P_i(t) \leq  0.001 \quad  \forall t \in \{1,2...,T\}\\
  &    
\end{aligned}
\end{equation}

\begin{figure}[h]
        \centering
        \begin{subfigure}[b]{0.4775\textwidth}
            \centering
        \includegraphics[width=8.5cm,height=5.592cm]{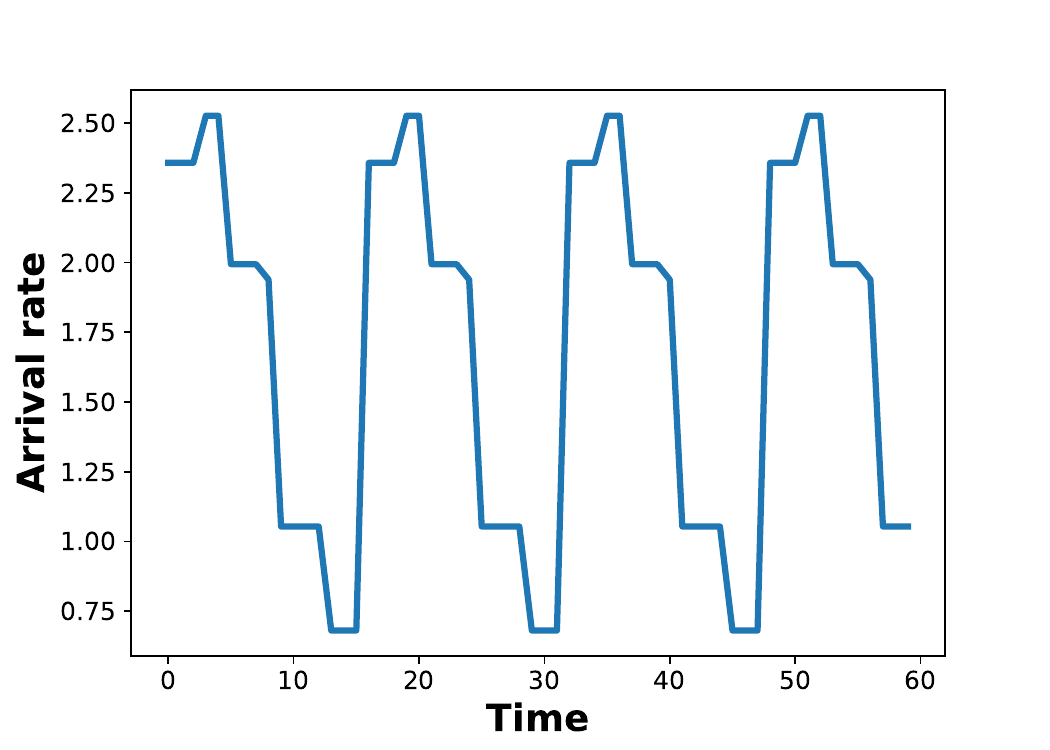}
            \caption[]%
            {{\small Arrival rates}}    
            \label{fig:arrival_rate_examples}
        \end{subfigure}
        \hfill
        \begin{subfigure}[b]{0.4375\textwidth}  
            \centering 
            \includegraphics[width=8.5cm,height=5.592cm]{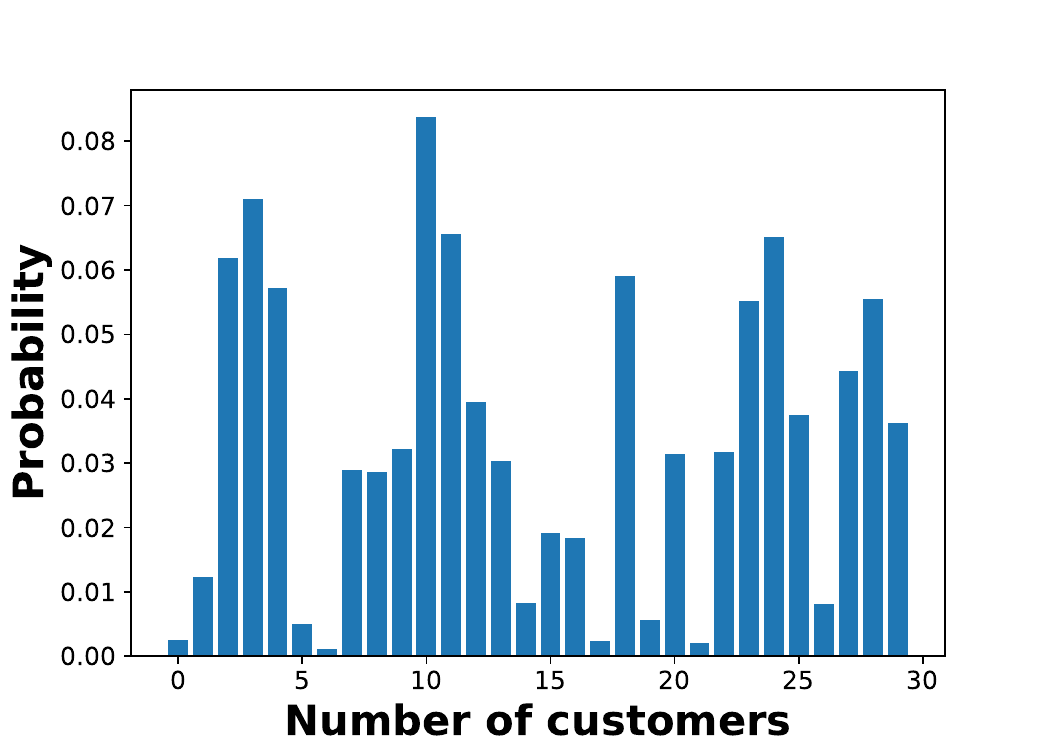}
            \caption[]%
            {{\small Initial distribution}}    
            \label{fig:initail_example}
        \end{subfigure}
        \caption{Input for the Optimization Example.}%
    \label{fig:test1_SAE_REM}
    \end{figure}

We consider an inter-arrival and service time distribution that follows an Erlang distribution with 2 phases. Since the service time follows an Erlang(2) distribution, the rate is changed via the rate parameter. As such, a rate change changes all distribution moments, not just the first. In the case of Erlang, the SCV remains constant. The arrival rates and the initial number of customers distribution are depicted in Figures~\ref{fig:arrival_rate_examples} and~\ref{fig:initail_example}. The feasible domain of service rate is $[0.1,10]$. We let $T=60$, $C_1 = 0.05$, and $C_2=10$.

\begin{figure}
\centering
\includegraphics[scale=0.42]{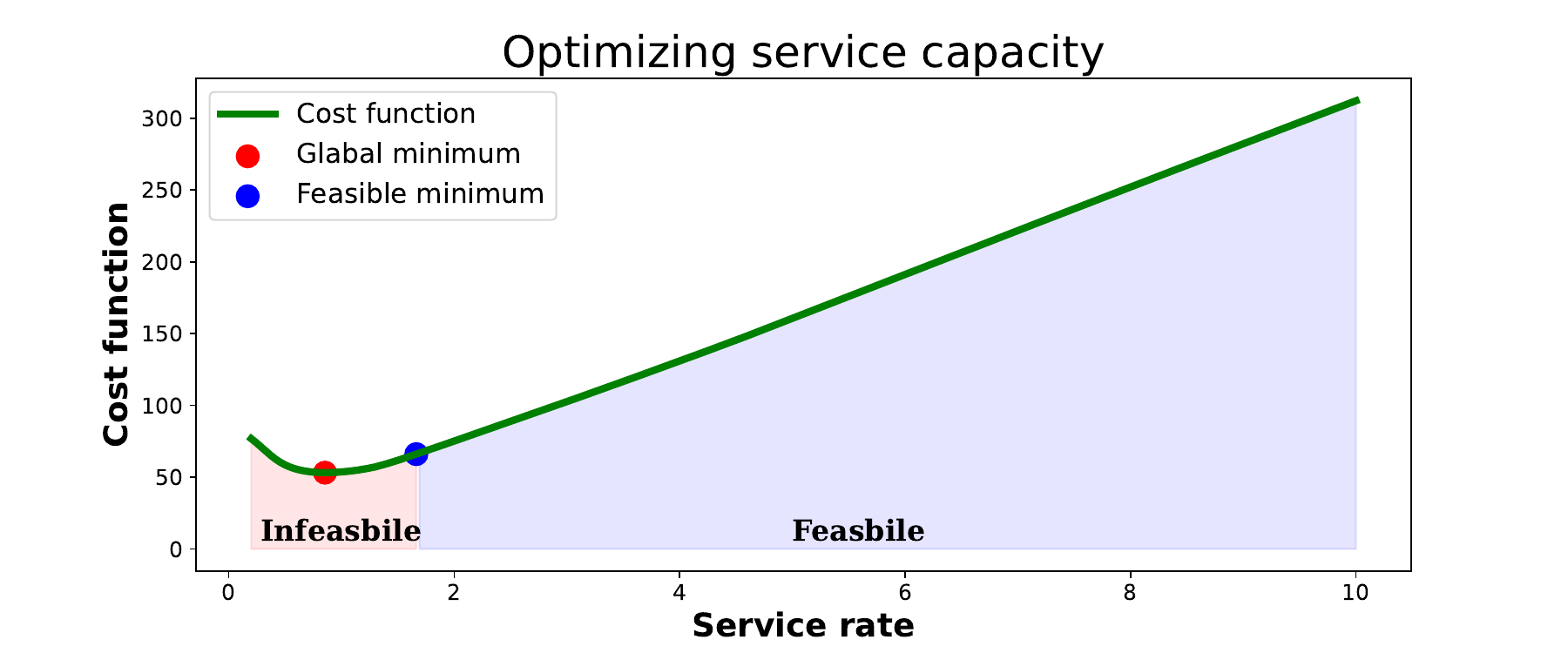}
\caption{Cost function.}
\label{fig:Cost_function}
\end{figure}

For optimization, we use a grid search over the entire domain of feasible service rates for 500 different values. As $\textit{MBRNN}$ can quickly infer many systems, we can easily apply this grid search here. The results are depicted in Figure~\ref{fig:Cost_function}. The solid line presents the cost function computed by the \textit{MBRNN} method. The graph also distinguishes between the infeasible and feasible areas of service rates concerning the constraint in Equaiton~\eqref{eq:constraint}. The red dot minimizes the cost at a capacity of 0.86 but violates this constraint; hence, the optimal capacity is 1.66, represented by the blue dot.

To validate the results, we simulate the system to check how accurate the \textit{MBRNN} was. Due to the long run times of simulations, we searched only around the service rate of 1.66. 
In total, we simulated four different service rate values (1.67,1.66,1.65,1.64). As the results in Table~\ref{tab:results_example} suggest, the service rate of 1.65 also meets the constraints and improves the 
cost function from 66.05 to 65.35, a savings of 1\%.  

The runtimes difference is significant; while a single simulation takes, on average, more than two and a half hours, 500 \textit{MBRNN} inferences take 0.015 seconds. Furthermore, if we were to use only simulation and therefore need to search the entire space, it would take even longer, making the optimization process via simulation impractical. For highly sensitive analysis, one can first use the \textit{MBRNN} and then simulate the system while focusing on smaller areas in the domain.

\begin{table}[htbp]
  \centering
  \caption{Results summary}
    \begin{tabular}{|c|c|c|c|c|}
    \toprule
          & Optimal service rate & Optimal cost function & Runtimes & \# of inferences \\
    \midrule
    MBRNN & 1.66  & 66.05 & 0.015 sec & 500 \\
    \midrule
    Simulation & 1.65  & 65.35 & 639.14 min & 4 \\
    \bottomrule
    \end{tabular}%
  \label{tab:results_example}%
\end{table}%

\subsection{Data-driven inference}\label{subsec:estimated_preds}

In this section, we discuss how to make inferences over a transient queueing system where the input, i.e., the inter-arrival process, service time distributions, and initial distribution, are not given. Instead, we have the event log of the queue, which also contains the information on the inter-arrival and service times, along with the initial number of customers at $t=0$. We illustrate this procedure with an example.  

For simplicity, we focus on the case where all arrivals originated from a single process. (Otherwise, one must decide when an arrival process ends, and the next begins, which requires further statistical tools such as clustering methods and change point analysis.) We take the following steps:

\begin{itemize}
    \item Sample a $GI/GI/1$ system based on Section~\ref{sec:data}, with a single arrival process. 
    \item Sample $n=50,000$ instances from the inter-arrival, service time, and initial state distributions. 
    \item Estimate the first four moments of the inter-arrival and service times using the standard unbiased estimator of the form $\bar{X^i} = \sum_{j=1}^{n} \frac{X^j}{n}, i=1, \ldots, 4$.
    \item  Estimate the initial distribution. We add the following notation:  Let $\hat{\textbf{P}}(0)_i$ be the estimated probability of having $i$ customers in the system at time $0$. Let $W_i$ be the frequency of the times the system had $i$ customers at time $t=0$ out of the $n$ samples. The estimate of $\hat{\textbf{P}}(0)_i$ is: 
   $$ \hat{\textbf{P}}(0)_i = \frac{W_i}{n}, \quad 0 \leq i \leq l.$$   
    This is the standard unbiased estimator. 
    \item Input the estimated inter-arrival and service time moments, together with the $\hat{\textbf{P}}(0)$ to the $\textit{MBRNN}$ model. 
\end{itemize}

\begin{figure}
\centering
\includegraphics[scale=0.42]{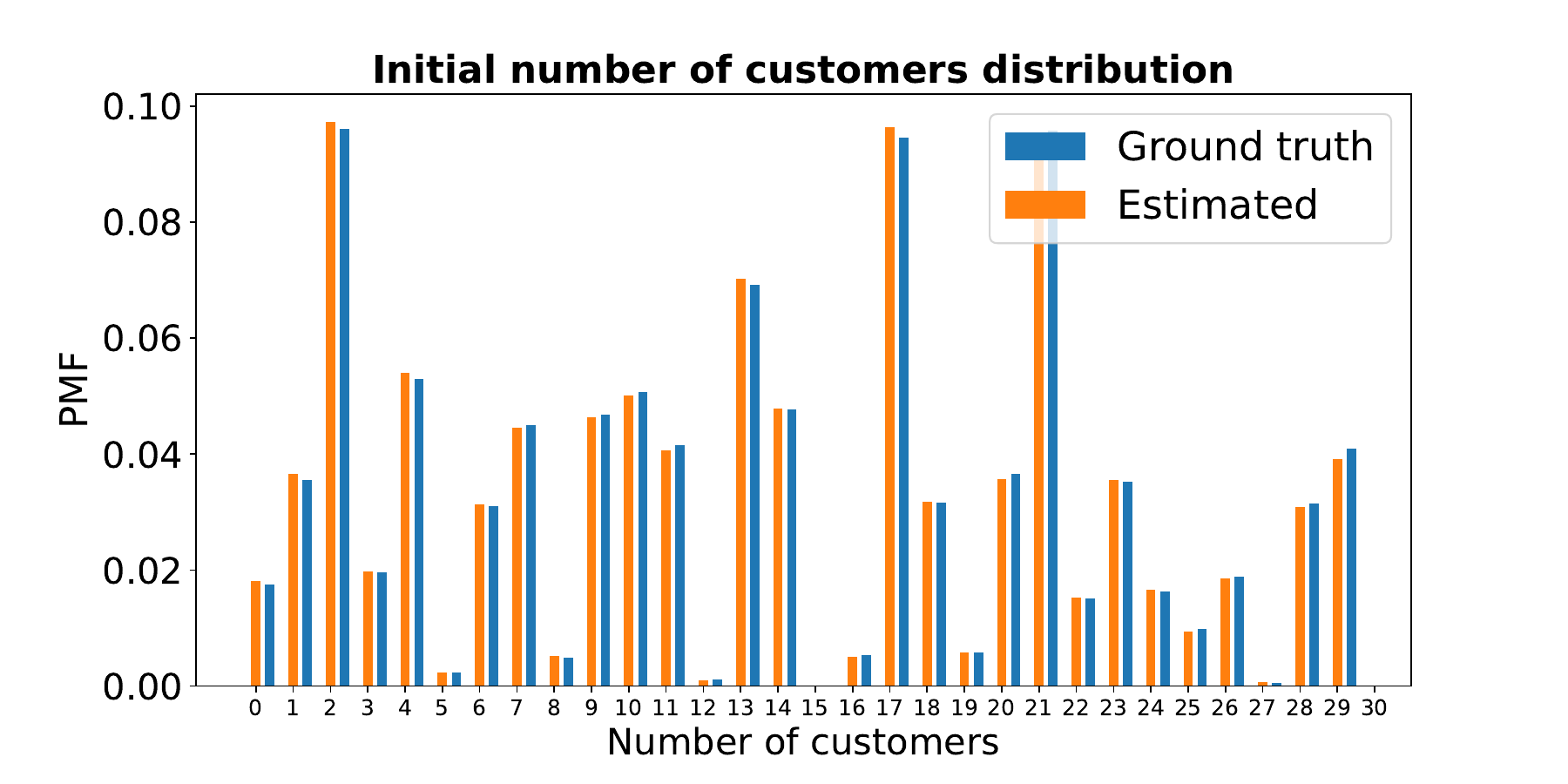}
\caption{Initial number of customer distribution for the Inference Example. }
\label{fig:data_driven_initial}
\end{figure}

In Table~\ref{tab:est_moms}, we present the true inter-arrival and service time moments (under "Ground truth") and the estimated moments (under "Estimated"). We also present the corresponding relative error of the estimate (under "Error")\footnote{We use the standard error: $100*(true-estimated)/true$}. As Table~\ref{tab:est_moms}, most of the errors are small, except the third and fourth of the inter-arrival moments, which exceed 3\% and 12\% error, respectively. In Figure~\ref{fig:data_driven_initial} we present the true initial number of customers distribution against the estimated one. The figure suggests a small error where the $\overline{SAE}$ distance between them is 0.017.

Feeding the \textit{MBRNN} with a slight error in the input should create some accuracy loss in the queue analysis. We present $\overline{SAE}$, $\overline{PARE}$, and $\overline{REM}$ measures (in comparison to the simulated "ground truth") in Table~\ref{tab:data_driven_acc}.

\begin{table}[htbp]
  \centering
  \caption{Estimated moments}
    \begin{tabular}{|c|c|c|c|c|c|c|}
    \toprule
          & \multicolumn{2}{c|}{Inter arrival moments} &       & \multicolumn{2}{c|}{Service time moments} &  \\
    \midrule
          & Ground truth & Estimated & Error \% & Ground truth & Estimated & Error \% \\
    \midrule
    1     & 1.69  & 1.69  & 0.42  & 1.00  & 1.00  & 0.02921 \\
    \midrule
    2     & 6.89  & 6.87  & 0.26  & 1.28  & 1.29  & 0.44886 \\
    \midrule
    3     & 48.22 & 46.46 & 3.77  & 2.19  & 2.21  & 1.10688 \\
    \midrule
    4     & 489.61 & 436.09 & 12.27 & 4.84  & 4.92  & 1.65866 \\
    \bottomrule
    \end{tabular}%
  \label{tab:est_moms}%
\end{table}%

The results suggest a small loss of accuracy in the $\overline{SAE}$. Strangely enough, the $\overline{PARE}$, the estimated input outperformed the system with the true input in inferring the $25^{th}$, $50^{th}$, $75^{th}$, and $90^{th}$ percentile.  In the REM, there was an accuracy decrease (from 0.53\% to 1.49\%) due to feeding the estimated input.  To conclude, estimating the input via data is not difficult, and for a sufficient amount of data, do not cause a significant error. 

\begin{table}[htbp]
  \centering
  \caption{\textit{MBRNN} accuracy}
    \begin{tabular}{|c|c|c|c|c|c|c|c|c|}
    \toprule
          &       & \multicolumn{6}{c|}{$\overline{PARE}$ }                    &  \\
    \midrule
          & $\overline{SAE}$   & 25    & 50    & 75    & 90    & 99    & 99.9  & $\overline{REM}$ \\
    \midrule
    True input & 0.027 & 2.08  & 1.52  & 1.15  & 0.94  & 0.95  & 0.94  & 0.53 \\
    \midrule
    Data-driven & 0.03  & 0.65  & 0.97  & 0.82  & 0.996 & 0.94  & 0.98  & 1.49 \\
    \bottomrule
    \end{tabular}%
  \label{tab:data_driven_acc}%
\end{table}%

\section{Discussion and Conclusion}\label{sec:Discussion}

We next discuss two aspects of the \textit{MBRNN} model. The first is an intuitive explanation of how the RNN model performs. The next is a possible extension of the method suggested here. 

\subsection{Intution of the \textit{MBRNN} model}\label{sec:intution}

The input of the \textit{MBRNN} model is the distribution of the number of customers in the system at $t=0$, the first four moments of both the inter-arrival and service time distributions. In addition, there is also the hidden input denoted by $H(\cdot)$, as in Figure~\ref{fig:NN_diagram}. It is (implicitly)  assumed that the inter-arrival and service time age is 0 at $t=0$. An intuitive way of understanding the model is that the hidden input $H(i)$ of the model keeps track of the number of customers in the system at the end of each period and the age (and hence the residual) of the inter-arrival and service time distributions. More broadly, $H(i)$ holds information from periods $i= \{1,2,...i-1\}$ that the MBRNN finds necessary for inference at future stages. 

Recall that the distribution of the number of customers at $t=0$ is provided as part of the input. We note that this input is applied at every period. As expected, this input is less important in later periods that rely greatly on the hidden input and more important in early periods that rely greatly on the initial state.    
\subsection{Extensions}\label{sec:extensions}

The \textit{MBRNN} predicts the transient number in the system distribution of $G(t)/GI/1$ system, where $G(t)$, i.e., the arrival process, is as described in Section~\ref{sec:arrival_proc}. Since the training data was generated via simulation and inference is done via a neural network, this method can be applied to more complex transient queueing systems. For example, allowing multiple servers, i.e., $G(t)/GI/c$ system, can be done by simply adding the number of servers to the input. The output will be kept the same as in this paper; if there are $c$ or fewer customers in the system, it implies all customers are in service.  

\begin{figure}
\centering
\includegraphics[scale=0.42]{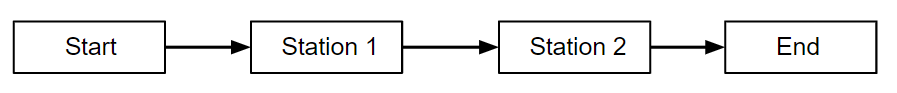}
\caption{Two Tandem queueing network.}
\label{fig:Two_system_tandem}
\end{figure}

Not only the extension to considering multiple servers can be handled under the \textit{MBRNN} framework. Another interesting, and yet possible, extension is a queueing network. Consider a two-queue tandem system as presented in Figure~\ref{fig:Two_system_tandem}. This system has an arrival process from the "Start" node, where all customers arrive at station 1. Then, they all proceed to station 2, leaving the system once they hit the "End"  node. If we were to be interested in the transient analysis of stations 1 and 2, the input would be the moments of the arrival process, the initial state distribution of both stations and the service moments of both stations.  The neural network should internally learn the arrival process of station 2 (i.e., the departure process of station 1) and store this information in the hidden input.  

Following this line of thought, one can extend our approach to more complex queueing systems. Furthermore, other aspects of the queueing system can be considered, such as reneging or abandonment, vacations, more complex arrival processes, etc.

We note that to create a rich, effective, but not too cumbersome training set, we can use known solutions (e.g., Jackson networks) and approximations (e.g., in heavy traffic).

\subsection{Conclusion}\label{sec:conclusions}

We use a neural network to analyze a single-station transient $G(t)/GI/1$ system. Our extensive performance evaluation suggests that our model is fast, accurate, and outperforms existing approximation methods. The inference procedure is simple and fast and relies only on the first four inter-arrival and service time moments and the initial state distribution. The \textit{MBRNN} model can make inferences in parallel for thousands of $G(t)/GI/1$  systems within a fraction of a second and is applicable for a greater number of periods than it was trained for. 

The \textit{MBRNN} model proposed here can serve as a prototype for a tool that analyzes real-life queueing systems. One can increase the labeling accuracy and training periods for software productization purposes. Furthermore, the  \textit{MBRNN} model can be extended to many transient queueing systems, including more complex queueing system architecture, time-varying behavior of arrivals and service times, and more. 

Creating an  \textit{MBRNN} tool for queueing analysis can be extremely useful. Most queueing systems in applications, and especially transient systems, are hard to analyze using a classic approach. Practically speaking, only simulations can solve such complex systems. However, as shown here, even after designing an effective simulation, its run time may be quite long, even for simple transient systems. Such long run times are detrimental to applications that require to evaluate the performance of thousands of systems, such as in system optimization. As the numerical example in Section~\ref{sec:numeric} indicates, our method produces similar results instantaneously. We, therefore, believe that 
\textit{MBRNN} is a vital tool for the modern queueing analyst.


\newpage

\backmatter









\bibliography{sn-article}

\begin{appendices}

\section{Test sets properties values}\label{append:SCV}

Here, we give more details regarding test sets 1 and 2.

\begin{figure}
\centering
\includegraphics[scale=0.82]{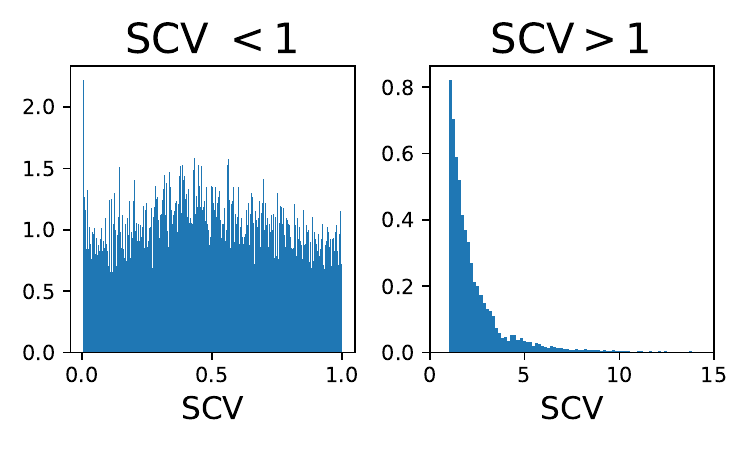}
\caption{SCV values test set 1. }
\label{fig:SCV_test_1}
\end{figure}

The SCV values used in test set 1 are in the range of $[0.004,394.2]$.  In Figure~\ref{fig:SCV_test_1}
we present the histogram of the SCV values of test set 1, separately for SCV values smaller than one and larger than one. 

\begin{figure}
\centering
\includegraphics[scale=0.82]{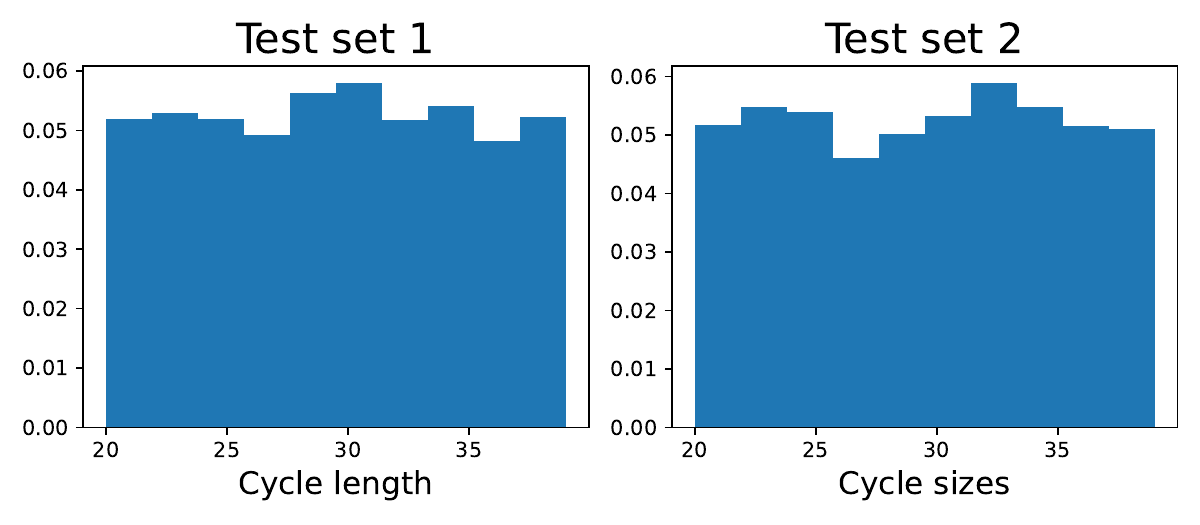}
\caption{SCV values test set 1. }
\label{fig:cycle_vals}
\end{figure}

The cycle length histograms of both test sets 1 and 2 are presented in Figure~\ref{fig:cycle_vals}. As depicted in the figure, both test sets are uniformly distributed according to our sampling procedure.  

\section{Distirubtions defintion}\label{append:dists_notation}

We introduce the notation for inter-arrival and service time distributions.

\begin{enumerate}
    \item Exponential ($M$) distribution with mean $1/\lambda$ SCV $c^2 =1$.
    \item Erlang ($E_k$) distribution with mean  $1/ \lambda$,  SCV $c^2 = 1/k$, i.e., the summation of k
i.i.d. exponential random variables, each with mean $1/(\lambda k)$.
    \item Hyperexponential $ (H_2(c^2))$  distribution, i.e., a mixture of two exponential distributions and exponential with mean $\lambda_1$ with probability $p_1$ and an exponential with mean $\lambda_2$ with probability $p_2=1-p_1$. The Hyperexponential distribution can be parameterized by its first three moments or the mean $1/\lambda$, SCV $c^2$ and
the ratio between the two components of the mean $r = p_1/\lambda_1 /(p_1/\lambda_1+p_2/\lambda_2)$ , where
$\lambda_1 > \lambda_2$. We only consider the case $r = 0.5$.
\item  Log-normal ($LN(c^2)$ ) distribution with mean 1 and scv $c^2$.
\item  Gamma ($G(4)$) distribution with mean 1 and scv $c^2 = 4$.
\end{enumerate}



\section{Fine-tuning of the NN}\label{append:Fine-tuning}

We provide technical details of the hyper-params fine-tuning of our model for a given ($n_{arrival}$, $n_{service}$). The hyper-params that we optimize are the learning rate, number of training periods, number of hidden layers, number of neurons for each layer, Batch-size, and the weight-decay parameter of the Adam optimizer (this is a regularization parameter; for more details on the Adam optimizer see~\cite{8624183} and reference therein). 

The state space for each parameter is:
\begin{itemize}
    \item Learning rate: $ \{0.0001, 0.0005, 0.001, 0.005\}$.
    \item Number of hidden layers: $\{2,3,4,5,6\}$.
    \item Number of neurons for each layer: $\{16,32,64,128\}$
    \item Batch-size: $\{32, 64, 128, 256\}$.
    \item Weight-decay: $\{10^{-4}, 10^{-5}, 10^{-6}\}$.
\end{itemize}
We do 500 random searches over these parameters and use the one with the lowest $\overline{SAE}$ over the validation set for each ($n_{arrival}$, $n_{service}$). Each search takes an average of 12.45 hours, training on an NVIDIA V100 Tensor Core GPU and 16 GB memory.
\end{appendices}


\end{document}